\pdfoutput=1

\documentclass[11pt]{article}

\usepackage[final]{acl}

\usepackage{times}
\usepackage{latexsym}
\usepackage[T1]{fontenc}
\usepackage[utf8]{inputenc}
\usepackage{microtype}
\usepackage{inconsolata}

\usepackage{booktabs}      
\usepackage{colortbl}      
\usepackage{pgfplots}
\usepackage[dvipsnames]{xcolor}
\usepackage{graphicx}      
\usepackage{amsmath}
\usepackage{amssymb}
\usepackage{amsfonts}
\usepackage{enumitem}
\usepackage[most]{tcolorbox}
\tcbuselibrary{skins, breakable, listings}
 \usepackage{tikz}
\usetikzlibrary{positioning, arrows.meta, calc}
\usepackage{xspace}
\usepackage[normalem]{ulem}
\usepackage{hyperref}
\usepackage{cleveref}
\Crefname{appendix}{Appendix}{Appendices}

\usepackage{subcaption}
\usepackage{fontawesome5}
\usepackage{makecell}

\usepackage{ellipsis} 
\usepackage{kotex}
\usepackage{multirow}

\pgfplotsset{compat=1.18}

\definecolor{highlight}{RGB}{50,102,173}
\definecolor{neutral}{RGB}{136,135,128}
\definecolor{axiscolor}{RGB}{136,135,128}

\newcommand{\CITE}[1]{{\color{green!80!black} [CITE]}}
\newcommand{\yes}{\checkmark}

\definecolor{hpered}{HTML}{D6604D}
\definecolor{seqblue}{HTML}{2166AC}

\newcommand{\APEseg}{\textsc{APE}\textsubscript{seg}\xspace}
\newcommand{\APErel}{\textsc{APE}\textsubscript{rel}\xspace}
\newcommand{\APEseq}{\textsc{APE}\textsubscript{seq}\xspace}
\newcommand{\APEfull}{\textsc{APE}\textsubscript{full}\xspace}
\newcommand{\APEexp}{\textsc{APE}\textsubscript{exp}\xspace}
\newcommand{\HPE}{{\footnotesize \faUserEdit}\,\textsc{HPE}\xspace}

\newcommand{\ddp}{\ensuremath{\mathsf{DDP}}\xspace}

\newcommand{\mGPT}{\textsc{\footnotesize GPT-5.4}\xspace}
\newcommand{\mGemini}{\textsc{\footnotesize Gemini-3-flash}\xspace}
\newcommand{\mGemma}{\textsc{\footnotesize Gemma-4-31B}\xspace}
\newcommand{\mHCX}{\textsc{\footnotesize HyperCLOVA-X-SEED-Think-32B}\xspace}
\newcommand{\mQwen}{\textsc{\footnotesize Qwen3.5-27B}\xspace}
\newcommand{\mDeepSeek}{\textsc{\footnotesize DeepSeek-V4-Pro}\xspace}

\newcommand{\mClaude}{\textsc{\footnotesize Claude Opus 4.6}\xspace}
\DeclareMathOperator{\ddpop}{ddp}

\newcommand{\xcomet}{\textsc{xComet}\xspace}
\newcommand{\xcometkiwi}{x\textsc{CometKiwi}\xspace}
\newcommand{\metricx}{\textsc{MetricX-24}\xspace}
\newcommand{\metricxqe}{\textsc{MetricX-24-QE}\xspace}
\newcommand{\dbleu}{d-\textsc{BLEU}\xspace}
\newcommand{\slide}{\textsc{SLIDE}\xspace}
\newcommand{\doccomet}{doc-\textsc{Comet}\xspace}
\newcommand{\ter}{\textsc{TER}\xspace}

\newtcolorbox{condbox}[1]{
    enhanced, 
    colback=gray!5,
    colframe=gray!20,
    coltitle=black,
    fonttitle=\bfseries,
    title=#1,
    left=5pt, 
    right=5pt, 
    top=2pt, 
    bottom=2pt,
    sharp corners,
    boxrule=0.5pt,
    underlay={
        \begin{tcbclipframe}
            \fill[gray!70] (frame.south west) rectangle ([xshift=3pt]frame.north west);
        \end{tcbclipframe}
    },
    before skip=5pt,
    after skip=5pt
}

\newtcolorbox{condboxA}[1]{
    enhanced, 
    colback=NavyBlue!3,
    colframe=NavyBlue!15,
    coltitle=black,
    fonttitle=\bfseries,
    title=#1,
    left=5pt, 
    right=5pt, 
    top=2pt, 
    bottom=2pt,
    sharp corners,
    boxrule=0.5pt,
    underlay={
        \begin{tcbclipframe}
            \fill[NavyBlue!60] (frame.south west) 
                rectangle 
                ([xshift=3pt]frame.north west);
        \end{tcbclipframe}
    },
    before skip=5pt,
    after skip=5pt
}

\newtcolorbox{condboxB}[1]{
    enhanced, 
    colback=OliveGreen!3,
    colframe=OliveGreen!15,
    coltitle=black,
    fonttitle=\bfseries,
    title=#1,
    left=5pt, 
    right=5pt, 
    top=2pt, 
    bottom=2pt,
    sharp corners,
    boxrule=0.5pt,
    underlay={
        \begin{tcbclipframe}
            \fill[OliveGreen!60] (frame.south west) 
                rectangle 
                ([xshift=3pt]frame.north west);
        \end{tcbclipframe}
    },
    before skip=5pt,
    after skip=5pt
}

\newcommand{\assumptionlabel}[3]{%
    \tcbox[
        colback=#1!12, 
        colframe=#1!40, 
        boxrule=0.4pt, 
        left=1pt, right=1pt, 
        top=0.5pt, bottom=0.5pt, 
        on line,
        arc=2pt
    ]{#3\,\textbf{\small #2}}%
}

\newcommand{\Aone}{\assumptionlabel{NavyBlue}{Size}{\small \faRulerHorizontal}}
\newcommand{\Atwo}{\assumptionlabel{OliveGreen}{Selection}{\small \faFilter}}

\newcommand{\circled}[1]{%
    \tikz[baseline=(char.base)]{
        \node[shape=circle, draw, inner sep=1pt, 
              font=\small\bfseries, line width=0.6pt, fill=gray!20] 
        (char) {#1};
    }%
}
\definecolor{LLMcolor}{HTML}{EE9B51}
\makeatletter
\newcommand\blfootnote[1]{%
  \begingroup
    \renewcommand\thefootnote{}%
    \let\orig@makefntext\@makefntext
    \def\@makefntext##1{\noindent##1}%
    \footnotetext{#1}%
    \addtocounter{footnote}{0}%
    \let\@makefntext\orig@makefntext
  \endgroup
}
\makeatother

\title{Discourse Dependency: A Continuous Criterion for Translation Difficulty}

\author{
  Ahrii Kim\textsuperscript{1} \quad
  Chanjun Park\textsuperscript{2 $\star$} \quad
  Seong-heum Kim\textsuperscript{1,3 $\star$} \\
  \textsuperscript{1}AI-Bio Convergence Research Inst. \quad
  \textsuperscript{2}School of Software \quad
  \textsuperscript{3}Dept. of Intelligent Semiconductors \quad
  \\ Soongsil University \\
  \texttt{\{ahriikim,chanjun.park,seongheum\}@ssu.ac.kr} \\
}

\begin{document}
\maketitle

\begingroup
\renewcommand{\thefootnote}{$\star$}
\footnotetext{Corresponding authors.}
\endgroup

\blfootnote{\faGithub{} \url{https://github.com/trotacodigos/ddp.git}}

\begin{abstract}

Recent calls for harder machine translation benchmarks have not clarified what difficulty should mean. We argue that one meaningful and currently unmeasured axis is \textit{referential reach}, the distance a segment must look back into its document to resolve the entities and pronouns it contains. We formalize this as \textit{\uline{d}iscourse \uline{d}e\uline{p}endency} (\ddp), a metric-free, source-side measure computed from named entity re-mentions and pronominal coreference. Validated against gold coreference, \ddp\ errs one-sidedly in $99.2\%$ of segments, so a high-\ddp\ segment is certified to require long-range context. Applying \ddp\ to WMT24++ and WMT25 shows that both are heavily skewed toward low-\ddp\ segments, which domain labels do not distinguish. Building on \ddp, we compare five context injection strategies in an English-Korean post-editing setup, varying context size and selection. As \ddp\ grows, no strategy keeps pace with human post-editing. On segments with $\ddp \geq 15$ raters prefer human translations, while automatic metrics register no difference. As frontier systems saturate aggregate scores, \ddp\ shifts evaluation from \textit{how well models score} to \textit{how far they can reach}.
\end{abstract}

\section{Introduction}
\label{sec:intro}

Translation quality has improved dramatically with large language models (LLMs), to the point where general-domain benchmarks can no longer reliably distinguish state-of-the-art systems \citep{kocmi-etal-2024-findings,kocmi-etal-2025-findings,akhtar2026aibenchmarksplateausystematic}. The machine translation (MT) community has responded by seeking harder evaluation data, either by selecting difficult instances from existing datasets or by generating them synthetically \citep{10.1162/TACL.a.60,proietti-etal-2025-estimating,zouhar-etal-2026-generating}. These efforts share a common premise: difficulty is estimated from translation quality scores, using automatic metrics as a proxy for how much a sentence challenges a model, a premise worth revisiting now that aggregate scores no longer differentiate frontier systems.

This premise has two limitations. First, automatic metrics are unreliable for the subtle quality differences that distinguish frontier systems \citep{mathur-etal-2020-tangled,freitag-etal-2022-results,lavie-etal-2025-findings}, so using them as difficulty estimators risks optimizing evaluation toward what metrics can measure rather than what genuinely challenges models. Second, the approach treats difficulty as a property of individual segments, while translation difficulty is often not intrinsic to a sentence but relational, depending on how far the information required to translate it faithfully lies within the document \citep{halliday1976cohesion,Hardmeier2012DiscourseIS}.

A parallel line of work has long recognized this relational nature of translation. Document-level MT studies show that discourse phenomena such as coreference, lexical cohesion, and tense consistency require information beyond the sentence \citep{voita-etal-2019-good, bawden-etal-2018-evaluating, fernandes-etal-2021-measuring}, and more recent efforts have moved toward systematic identification of context-dependent segments through automatic taggers or rule-based extraction pipelines \citep{fernandes-etal-2023-translation, wicks-post-2023-identifying}. These evaluations remain phenomenon-specific: a sentence either falls under a target phenomenon or it does not, and coverage depends on per-language, per-phenomenon rule engineering. What is missing is a continuous, document-level measure that quantifies how much contextual reach a segment requires, applicable uniformly across genres and datasets to characterize a benchmark's overall difficulty profile.

We therefore reframe difficulty around a single document-level property, \textbf{\uline{d}iscourse \uline{d}e\uline{p}endency (\ddp)}: the distance within a document at which discourse-relevant information appears relative to the current segment, operationalized as the maximum distance between successive mentions of the same entity. \ddp is metric-free and source-side, which lets it serve as a difficulty criterion independent of any model or scoring function. It also recasts the role of domain in MT evaluation: rather than a topical label, domain becomes a coarse proxy for the discourse dependency that \ddp\ measures directly.

To test whether \ddp\ tracks translation difficulty, we examine how models exploit context in automatic post-editing (APE) for English-Korean (En-Ko), a language pair known to be challenging both linguistically and for translation evaluation \citep{park-pado-2024-multi, choi-etal-2018-automatic, lee-etal-2025-testset}. Our main contributions are as follows:
\begin{itemize}[noitemsep]
    \item \textbf{Discourse dependency (\ddp).} A linguistically grounded, metric-free difficulty criterion computed from entity re-mentions and pronominal coreference, exposing within-domain variation that domain labels conflate.

    \item \textbf{A \ddp-stratified evaluation framework.} Five context injection strategies paired with \ddp-stratified analysis, providing a discourse-level account of when context utilization breaks down.

    \item \textbf{A diagnostic of automatic metrics.} \ddp\ stratification exposes a blind spot in standard metrics, providing a concrete axis for auditing metric reliability beyond aggregate scores.

    \item \textbf{\ddp-annotated benchmarks.} \ddp\ statistics for WMT24++ and WMT25, showing that current evaluation data is heavily skewed toward low-\ddp\ segments.
\end{itemize}

Organizing evaluation along this axis supports test sets stratified by discourse progression rather than instance-level difficulty selection, targeting the inter-sentential reasoning that current systems still lack. It also exposes how much within-domain variation topical labels conceal (\Cref{subsec:wmt_analysis}).

\section{Related Work}

\paragraph{Difficulty in MT evaluation.}
As frontier MT systems converge on general-domain benchmarks, recent work has shifted toward harder evaluation data \citep{kocmi-etal-2024-findings, kocmi-etal-2025-findings}. One line selects difficult instances from existing corpora using metric-based filtering or model disagreement \citep{10.1162/TACL.a.60, proietti-etal-2025-estimating}, while another generates challenging instances synthetically, often through perturbation or targeted construction \citep{zouhar-etal-2026-generating}. Both rely on automatic quality scores to define difficulty, inheriting the known limitations of these metrics at the high end of the quality scale \citep{mathur-etal-2020-tangled, freitag-etal-2022-results,lavie-etal-2025-findings}. \ddp\ departs from this framing by deriving difficulty from a source-side, document-level discourse property, independent of any model or metric. Metric-derived estimates require system outputs, so they can filter an existing test set but cannot inform its design, whereas \ddp\ is available before any translation exists. We therefore do not expect the two to correlate, since a source-side criterion that reproduced quality scores would add nothing (\Cref{sec:results}).

\paragraph{Discourse phenomena in MT evaluation.}
Sentence-level evaluation obscures discourse-sensitive errors, prompting two complementary responses. WMT human evaluation has progressively expanded rater context from neighboring sentences to paragraph-level pilots \citep{barrault-etal-2019-findings, akhbardeh-etal-2021-findings, kocmi-etal-2022-findings, kocmi-etal-2023-findings}, though ratings still tend toward ceiling effects \citep{kocmi-etal-2024-findings}. In parallel, contrastive challenge sets and WMT test suites isolate phenomena such as pronoun resolution, lexical cohesion, and ellipsis through binary preference on minimal pairs \citep{muller-etal-2018-large, voita-etal-2019-good, manakhimova-etal-2024-investigating, bhattacharjee-etal-2024-domain-dynamics}, while recent work moves toward systematic identification of context-dependent segments \citep{fernandes-etal-2023-translation, wicks-post-2023-identifying} and toward measuring whether models actually use the supplied context \citep{mohammed-niculae-2024-measuring}. A shared finding across these lines is that standard WMT test sets contain few context-dependent sentences, with document-level gains visible only on discourse-dense subsets \citep{post2024escapingsentencelevelparadigmmachine}. Yet discourse remains operationalized through phenomenon-specific annotations or binary classifications, leaving open the question of \textit{how much} contextual reach a segment requires.

\paragraph{Context-aware MT.}
Approaches to incorporating document context have evolved along two axes: how much context to provide, and how to select it. Early neural systems concatenate a fixed window of $k$ preceding segments \citep{tiedemann-scherrer-2017-neural, junczys-dowmunt-2019-microsoft}, with hierarchical and cache-based variants encoding surrounding sentences through separate modules \citep{miculicich-etal-2018-document, maruf-etal-2019-selective}. This sequential window remains the de facto standard, though architectural extensions yield diminishing returns once context is long enough \citep{sun-etal-2022-rethinking, post2024escapingsentencelevelparadigmmachine}. The shift to LLM-based MT has reframed the problem from architecture to prompting. Frontier LLMs can ingest full documents directly \citep{wang-etal-2023-document-level, karpinska-iyyer-2023-large}, yet they do not exploit extended input uniformly and often degrade well before the nominal limit \citep{liu-etal-2024-lost, mohammed-niculae-2024-measuring}. Our results offer a discourse-level account of this failure. The context utilization degrades systematically as \ddp\ grows, indicating that the bottleneck is not input length but the linguistic reach a segment demands.
\section{\ddp: Discourse Dependency}

\subsection{Theoretical Background}
\label{subsec:concept}
The difficulty of translating a segment depends not only on its internal structure but on where in the document its required information lies. A linguistically simple segment can be contextually demanding when its referents are introduced many segments earlier. Consider the following passage:

\begin{quote}
\textit{(1) \textcolor{green!70!black}{John} arrived.\\
(2) \textcolor{blue}{He} looked tired.\\
(3) \textcolor{green!70!black}{The manager} greeted \textcolor{blue}{him}.\\
(4) \textcolor{green!70!black}{John} smiled.\\
(5) \textcolor{blue}{She} offered \textcolor{blue}{him} coffee.}
\end{quote}

\noindent Translating segment (5) requires knowing that \textit{she} refers to \textit{the manager} from (3), and that \textit{him} refers to \textit{John} from (1). A system that sees only the immediately preceding segment cannot resolve either reference.

We formalize this intuition as \textit{\uline{d}iscourse \uline{d}e\uline{p}endency} (\ddp). Each segment contains \textit{anchors}, entity mentions whose interpretation depends on a prior segment \citep{HOBBS1978311, 10.5555/12457.12458}, and \ddp\ records the distance from each anchor to its antecedent according to two linguistically motivated rules: \circled{1} \textbf{a named entity is usually the first mention in the document}, since it typically introduces new information. If a prior mention exists, the distance is measured to it \citep{Prince1981TowardAT}. \circled{2} \textbf{A pronoun is anchored to the most recent named entity or head noun of compatible gender and number}, following the accessibility hierarchy of referential expressions \citep{clark1977comprehension, grosz-etal-1995-centering}. The segment-level \ddp\ is the maximum anchor distance within the segment. \Cref{tab:ddp_example} illustrates the computation on the passage above. \ddp\ measures one axis, referential reach, and is a \textit{lower bound} on contextual dependency rather than a general model of translation
difficulty (\Cref{subsec:operation}).

\begin{table}[t]
\small
\centering
\begin{tabular}{llll}
\toprule
\textbf{Seg} & \textbf{Anchor} & 
\textbf{Antecedent} & \textbf{$d$} \\
\midrule
(1) & \textit{John}    & --- (first mention) & 0 \\
(2) & \textit{he}      & \textit{John} (1)   & 1 \\
\midrule
(3) & \textit{manager} & --- (first mention) & 0 \\
    & \textit{him}     & \textit{John} (1)   & 2 \\

    & \multicolumn{2}{r}{$\ddpop_{(3)} = \max(0, 2)$} & \textbf{2} \\
\midrule
(4) & \textit{John}    & \textit{John} (1)   & 3 \\   
\midrule
(5) & \textit{she}     & \textit{manager} (3)& 2 \\
    & \textit{him}     & \textit{John} (1)$^\dagger$ & 4 \\
    & \multicolumn{2}{r}{$\ddpop_{(5)} = \max(2, 4)$} 
    & \textbf{4} \\
\bottomrule
\end{tabular}
\caption{\ddp computation on the example passage. Each anchor is traced to its antecedent; the segment \ddp is the maximum anchor distance ($d$). $^\dagger$\textit{him} is chained through its most recent named entity (\textit{John}, seg 4), whose own distance from its first mention (seg 1) is accumulated: $(5{-}4) + 3 = 4$.}
\label{tab:ddp_example}
\end{table}

\subsection{Operationalization}
\label{subsec:operation}

\paragraph{Anchor extraction.}
We extract two types of discourse anchors from the English source: named entities and referential pronouns.

\textit{Named entities} are identified by a named entity recognizer (NER), restricted to four categories — \textsc{person}, \textsc{org}, \textsc{gpe}, \textsc{loc} — which are the types most likely to persist across segments and require discourse-level resolution.

\textit{Pronouns} are identified through part-of-speech (POS) tagging (\texttt{pos\_ == PRON}) and restricted to referential uses, and expletive pronouns (e.g., \textit{it} in \textit{it is raining}) are excluded via dependency parsing (\texttt{dep\_ == expl}). Pronouns are grouped by gender and number into four classes: \textsc{masc} (\textit{he, him, his}), \textsc{fem} (\textit{she, her, hers}), \textsc{neut} (\textit{it, its}), and \textsc{plur} (\textit{they, them, their}).\footnote{Singular gender-neutral \textit{they} (e.g., for non-binary referents) is not separately classified and falls into the \textsc{plur} group.} Head nouns in subject or object position (\texttt{dep\_ $\in$ \{nsubj, dobj, nsubjpass\}}, \texttt{pos\_ == NOUN}) that are not captured by NER serve as fallback antecedents.

\begin{figure*}[ht]
    \centering
    \includegraphics[width=1\linewidth]{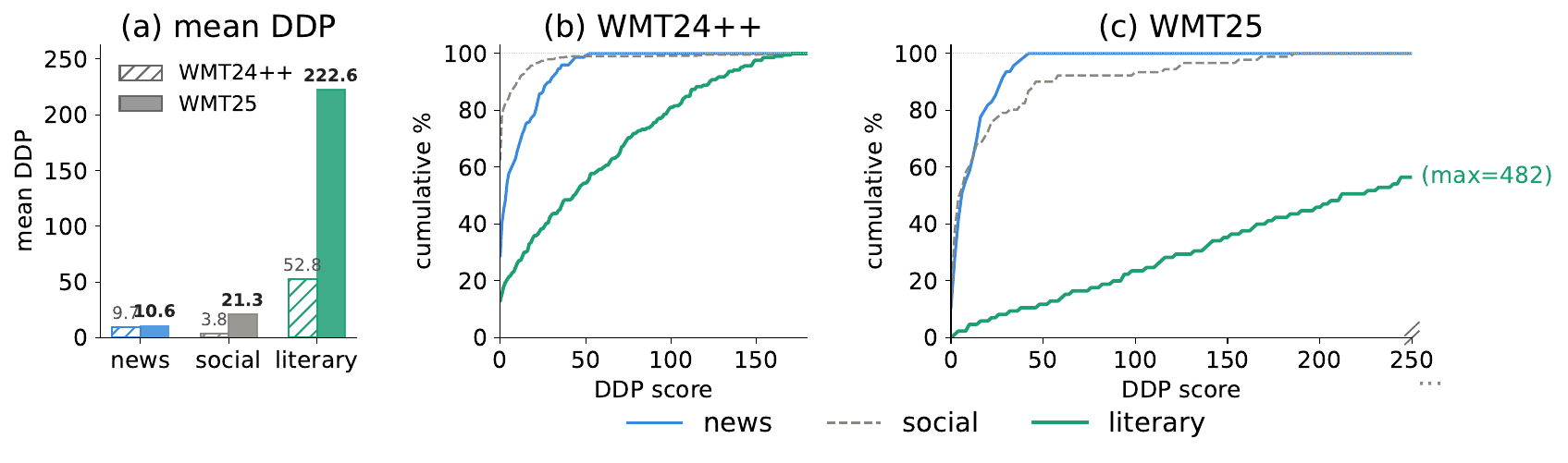}
    \caption{
        \ddp\ score distributions across domains in WMT24++ and WMT25. \textbf{(a)} Mean \ddp\ per domain: literary texts exhibit substantially higher dependency than news and social across both benchmarks, with WMT25 showing a marked increase across all domains ($\mu$=222.6 for literary vs.\ 52.8 in WMT24++) due to the inclusion of long-form narrative texts and longer conversational threads. \textbf{(b--c)} Cumulative distribution of \ddp\ scores: news segments saturate quickly in both benchmarks, while literary scores spread widely, reaching $\ddpop=171$ in WMT24++ and $\ddpop=482$ in WMT25. Note the different x-axis ranges in (b) and (c).
    }
    \label{fig:ddp_dist}
\end{figure*}

\paragraph{Distance computation.}
We process each document sentence by sentence, aligning the computation with the conventional evaluation unit. For each anchor in sentence $s_i$, we compute a distance $d(s_i, a)$ according to Rules \circled{1} and \circled{2}.
\vspace{0.5em}

\noindent\textbf{Rule \circled{1} (named entities).} Let $e$ be a named entity mention in $s_i$, and let $s_j$ ($j < i$) be the earliest prior sentence containing a mention of the same entity:
\begin{equation}
    d(s_i, e) = 
    \begin{cases}
        0 & \text{if no prior mention exists,} \\
        i - j & \text{otherwise.}
    \end{cases}
\end{equation}
 
\noindent\textbf{Rule \circled{2} (pronouns).} A pronoun $p$ in $s_i$ of gender-number group $g$ cannot be a first mention, so it is anchored to the most recent named entity or head noun of compatible group $g$ in some prior sentence $s_k$ ($k < i$).
\begin{equation}
    d(s_i, p) =
    \begin{cases}
        (i - k) + d(s_k, e_k) & \text{named entity } e_k,\\
        i - k                 & \text{head noun,}\\
        1                     & \text{no antecedent.}
    \end{cases}
\end{equation}

\noindent When the antecedent is a named entity, the chain accumulates its own distance $d(s_k, e_k)$ from its first mention. A head noun outside NER carries no such distance, so no accumulation applies. When no compatible antecedent exists, $d = 1$ reflects a minimal dependency on the immediately preceding sentence.

Each anchor in $s_i$ is computed independently, and the sentence-level \ddp\ is the maximum distance across all anchors $\mathcal{A}_i$:
\begin{equation}
    \ddpop(s_i) = 
    \max_{a \in \mathcal{A}_i} d(s_i, a).
    \label{eq:ddp}
\end{equation}

\noindent Sentences with no anchors, or only first-mention anchors, receive $ddp = 0$, indicating that no prior context is required for their translation.

\paragraph{Evaluation unit.}
Distances are counted in sentences and searched over the whole document, so an antecedent may lie anywhere before $s_i$ regardless of segment boundaries. Segments are only the unit at which we report. When a segment spans several sentences we take the maximum,
\begin{equation}
    \ddpop(\mathrm{seg}) =
    \max_{s_i \in \mathrm{seg}} \ddpop(s_i).
    \label{eq:ddp_seg}
\end{equation}
\noindent Segmentation therefore affects only how the sentence-level distances are grouped for reporting, not the distances themselves.

\paragraph{Implementation.}
NER, POS tagging, dependency parsing, and sentence segmentation are performed jointly with spaCy \texttt{en\_core\_web\_trf}, a transformer-based pipeline (RoBERTa backbone) that achieves state-of-the-art accuracy on OntoNotes \citep{hovy-etal-2006-ontonotes}. \Cref{appx:ablation} confirms that \ddp\ is robust to the choice of NER and POS tagger. Replicating the full pipeline with the architecturally distinct Flair tagger \citep{akbik-etal-2019-flair} reproduces segment-level \ddp\ at Pearson $r=0.89$ with $69\%$ exact agreement, and preserves the domain ordering. All computations run on the English source, making \ddp\ independent of the target language and translation system.

\definecolor{KoMark}{HTML}{1F77B4}
\newcommand{\komark}{\textcolor{KoMark}{\raisebox{0.7ex}{\scriptsize$\blacklozenge$}}\,}

\begin{table*}
\centering
\small
\resizebox{\textwidth}{!}{
\begin{tabular}{lccccccc}
\toprule
\textbf{Model} & \textbf{Params} & \textbf{Context} & \textbf{Cutoff} & \textbf{En-X} & \textbf{Reasoning} & \textbf{MoE} & \textbf{Thinking} \\
\midrule
\rowcolor{gray!15}
\multicolumn{8}{l}{\textbf{Open-source}} \\
\mGemma \citep{gemma4} & 30.7B & 256K$\rightarrow$128K & Jan 2025& \yes & \yes & -- & \yes \\
\mQwen \citep{qwen3.5} & 27B & 256K$\rightarrow$128K & -- & \yes & -- & -- & \yes \\
\mHCX \citep{navercloudhyperclovaxteam2026hyperclovax32bthink}\komark & 32B & 128K & -- & \yes & \yes & -- & \yes \\
\mDeepSeek \citep{deepseekai2026deepseekv4} & 1.6T {\footnotesize (A49B)} & 1M$\rightarrow$128K & -- & -- & \yes & \yes & \yes \\
\midrule
\rowcolor{gray!15}
\multicolumn{8}{l}{\textbf{Closed}} \\
\mGPT \citep{singh2026openaigpt5card} & -- & 1.05M & Aug 2025 & \yes & \yes & -- & \yes \\
\mGemini \citep{google2026gemini3flash} & -- & 1M & Jan 2025 & \yes & \yes & -- & \yes \\
\bottomrule
\end{tabular}
}
\caption{LLMs evaluated across input configurations. \textit{Cutoff} indicates the knowledge cutoff date as publicly disclosed; ``--'' denotes undisclosed. \textit{En-X} indicates whether the model was explicitly trained on X$\in$ \{Ko, Zh\}. \textit{Reasoning} = general reasoning capability, \textit{MoE} = Mixture of Experts architecture, \textit{Thinking} = a chain-of-thought mode is \emph{available}. It is disabled in all our runs for comparability (\Cref{sec:setup}). \komark~Korean-specialized models. Context lengths are restricted to 128K.}
\label{tab:model_list}
\end{table*}

\paragraph{Scope and validity as a lower bound.}
Unmodeled phenomena such as lexical cohesion and bridging anaphora, and links missed by the tagger, lower \ddp\ but do not inflate it, since inflation would require a spurious \emph{early} entity mention. The guarantee is thereforeone-sided. A high-\ddp\ segment is certified to require long-range context, while a low-\ddp\ segment is \emph{not} certified to be easy, and every claim in this paper rests on that asymmetry. To check that the bound is tight enough to be useful, we validate \ddp\ against gold coreference annotations from OntoNotes 5.0 on 80 documents (\Cref{app:ontonotes}). Segment-level \ddp\ correlates with gold distance at Pearson $r = 0.81$, and the error is one-sided in $99.2\%$ of segments, with the residual $0.8\%$ attributable to NER false positives.

\subsection{WMT Sets Through the Lens of \ddp}
\label{subsec:wmt_analysis}
To show that \ddp\ captures variation that domain labels miss, we analyze two widely used MT benchmarks: WMT24++ \citep{deutsch-etal-2025-wmt24} and WMT25 \citep{kocmi-etal-2025-findings}. \Cref{fig:ddp_dist} shows the segment-level \ddp\ distribution across domains, and \Cref{tab:ddp_stats} reports summary statistics.

Across both benchmarks, literary texts exhibit substantially higher \ddp\ than news and social domains (\Cref{fig:ddp_dist}a). In WMT24++, literary segments have a mean $\ddpop=52.8$ ($\sigma=47.0$), compared to 9.7 for news and 3.8 for social. WMT25 follows the same ordering, with literary reaching a mean of 222.6 (max 482) due to long-form narratives with character references spanning hundreds of sentences. WMT25 also shifts social upward to $\mu=21.3$, reflecting longer conversational threads than in WMT24++. Within WMT24++, social falls below news ($\mu=3.8$ vs.\ 9.7), against the assumption that conversational texts carry rich discourse structure. \textit{Discourse dependency thus varies within domains as much as across them}: a news article with persistent entity references can be harder to translate than a literary passage of self-contained sentences, yet both carry the same domain label.

The cumulative distributions (\Cref{fig:ddp_dist}b--c) reveal a structural bias in both benchmarks: news segments saturate well below $\ddpop=50$, while literary segments extend far into the high-dependency range. WMT25 covers substantially more high-\ddp\ texts than WMT24++, yet the bulk of segments in both remain at low to moderate dependency. Current MT benchmarks thus systematically undersample contextually difficult cases, providing quantitative grounding for \citet{post2024escapingsentencelevelparadigmmachine}'s observation that document-level gains surface mainly on discourse-dense subsets. This limitation becomes critical when evaluating document-level systems.
\section{Experimental Setup}
\label{sec:setup}

\subsection{Data}
We use WMT24++ \citep{deutsch-etal-2025-wmt24}, which provides professionally post-edited translations (\HPE) across four domains. We focus on English-Korean (En-Ko), a target language particularly sensitive to discourse. Korean's pro-drop structure requires inter-sentential restoration of omitted subjects and objects \citep{lee-etal-2025-testset}, and its honorific and named-entity systems demand cross-segment consistency, making En-Ko challenging for both translation and evaluation \citep{park-pado-2024-multi, choi-etal-2018-automatic}. After excluding the speech domain and documents with fewer than two segments, we retain 59 En-Ko documents spanning literary, news, and social. Dataset statistics are in \Cref{appx:data_stat}. We also report English-Chinese (En-Zh) results in \Cref{appx:enzh} for cross-lingual consistency.

\paragraph{Why post-editing.}
We evaluate in an APE setup, in which a model revises an existing draft, rather than translating from scratch. This isolates our research question, whether models \emph{use} the context they are given, from raw generation quality, which would otherwise confound not-using-context with not-translating-well. It also reduces the benefit of memorized references (\nameref{sec:limit}). Since \ddp\ is computed source-side, the stratification itself is unaffected by this choice.

\subsection{Models}
We employ six state-of-the-art LLMs spanning open-weight and proprietary systems, selected to represent diversity in parameter scale, architecture, and language specialization (\Cref{tab:model_list}). All models are prompted with a uniform 128K context limit, thinking mode disabled for comparability (the \textit{Thinking} column of \Cref{tab:model_list} records whether a chain-of-thought mode is \emph{available}, not whether it is used), and fixed decoding parameters (\texttt{temperature=1.0}, \texttt{top\_p=0.95}, \texttt{seed=42}), following recent LLM-based MT evaluation protocols \citep{kocmi-etal-2025-findings}.

\subsection{Context Injection Strategies}
\label{subsec:strategies}

\begin{condboxA}{\APEseg \hfill \Aone}
{\footnotesize No context is provided. Null baseline against all context-aware strategies.}
\end{condboxA}

\begin{condboxA}{\APEfull \hfill \Aone}
{\footnotesize The entire document, excluding the current segment pair. Maximizes context quantity while preserving sequential structure.}
\end{condboxA}

\begin{condbox}{\APEseq \hfill \Aone\ \Atwo}
{\footnotesize The $k$ segment pairs immediately preceding $s_i$, replicating the de facto standard in prior APE work.}
\end{condbox}

\begin{condboxB}{\APErel \hfill \Atwo}
{\footnotesize The $k$ segment pairs most semantically similar to $s_i$, retrieved via embedding-based search.}
\end{condboxB}

\begin{condboxB}{\APEexp \hfill \Atwo}
{\footnotesize Structured declarative knowledge of genre, participant relationships, and register, extracted from the source document by prompting an LLM.}
\end{condboxB}

If \ddp captures genuine contextual difficulty, models should benefit more from broader context as \ddp grows, where the information required for faithful translation lies beyond the immediate window. We test this by varying \textit{context injection}, the deliberate selection and structuring of information placed in the context window for post-editing. Given a document of $n$ segments, let $s_i$ denote the $i$-th segment under APE and $k$ the number of context segments provided. We design five strategies organized around two assumptions, marked in the boxes above: \textbf{\Aone}, that more context is better, and \textbf{\Atwo}, that proximity implies relevance. \APEseq\ serves as the shared pivot against which both are tested.

\Aone\ is tested by comparing \APEseg, \APEseq, and \APEfull, which provide zero, $k$, and full-document context. \Atwo\ is tested by comparing \APEseq, \APErel, and \APEexp, which fix context size ($k{=}5$) and vary the selection criterion. Prompt structure is held identical across strategies, with only the context block varying (\Cref{appx:prompt}), and construction details are in \Cref{appx:ctx_construct}.

\subsection{Evaluation Protocol}

\paragraph{Human ranking.}
\label{para:hum_eval}
We conduct relative ranking on two non-overlapping subsets of ${\approx}$500 segments, one per assumption. Annotators rank four candidate translations (including \HPE) from 1 to 4 with ties allowed. Each segment receives two independent judgments. For analysis, ranks are reversed to preference scores on a 1--4 scale, so that higher values indicate stronger preference. Tied candidates receive the average of the ranks they span, so that the four scores always sum to 10. Inter-annotator agreement measured by Kendall's $\tau$ is 0.37 on average, consistent with prior MT ranking studies \citep{callison-burch-etal-2007-meta}. Details are in \Cref{appx:humeval}.

\begin{figure*}
    \centering
    \includegraphics[width=1\linewidth]{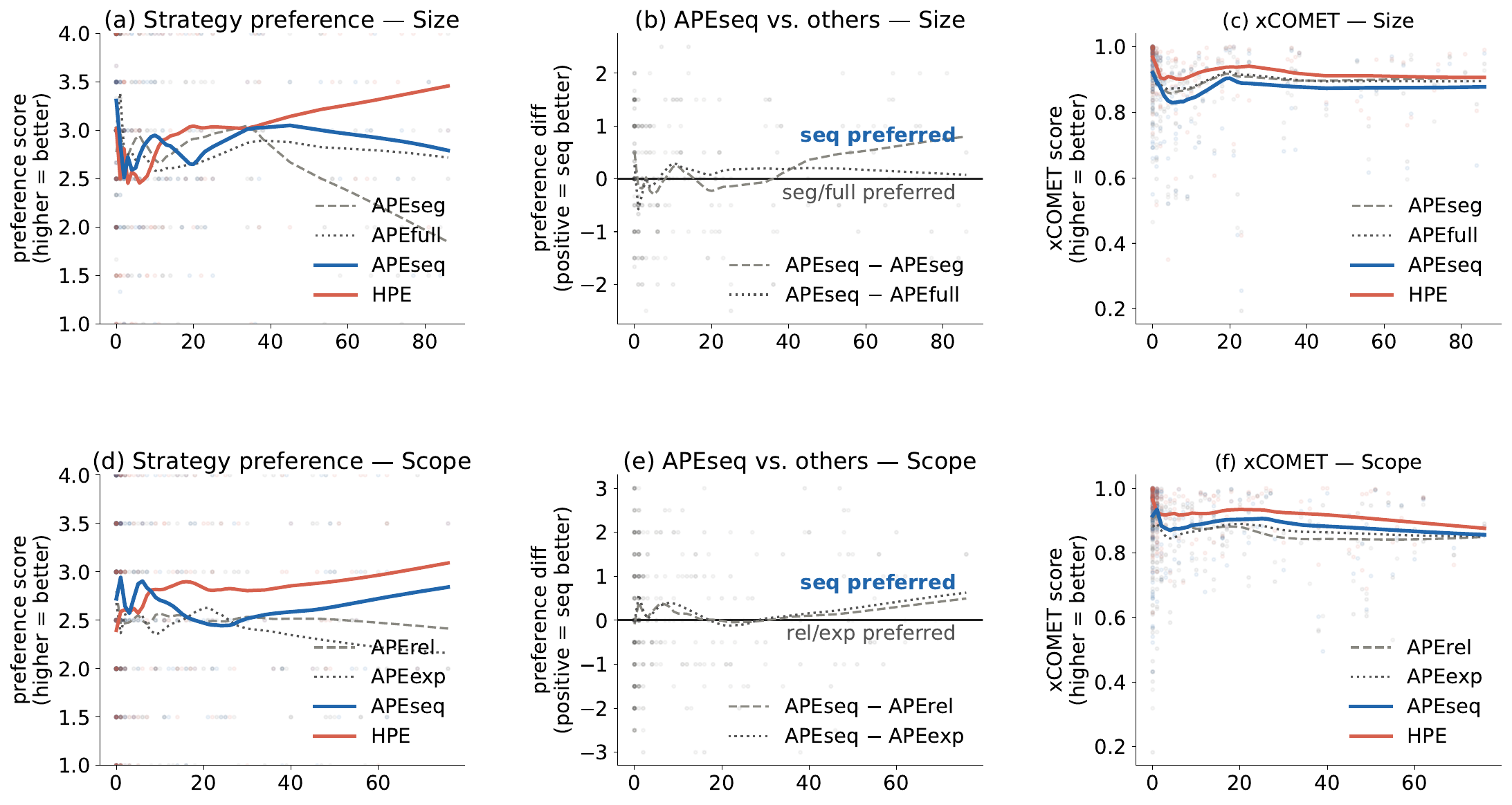}
    \caption{LOWESS-smoothed preference scores and pairwise differences as a function of segment-level \ddp. \textbf{Left panels} show the mean preference score (higher = better) of each strategy across the \ddp\ spectrum. \textbf{Middle panels} show the preference difference between \APEseq\ and each competing strategy (positive = \APEseq\ preferred). \textbf{Right panels} show the corresponding \xcomet scores for comparison with human judgments. 
    }
    \label{fig:ddp_lowess}
\end{figure*}

\paragraph{Automatic metrics.}
We report \ter \citep{snover-etal-2006-study} to quantify edit size, four segment-level metrics (\xcomet-XXL: \citealt{10.1162/tacl_a_00683}, \xcometkiwi-XXL: \citealt{rei-etal-2023-scaling}, \metricx-XXL, and \metricxqe-XXL: \citealt{juraska-etal-2024-metricx}), and three document-level metrics (\dbleu: \citealt{10.1162/tacl_a_00343}, \slide: \citealt{raunak-etal-2023-evaluating}, \doccomet: \citealt{vernikos-etal-2022-embarrassingly}) for translation quality. Outputs with \ter~$>$~100 are excluded as generation failures \citep{raunak:23-leveraging}.

\paragraph{\ddp-stratified analysis.}
For each segment, we compute its \ddp\ and the mean preference score per strategy, converted from rank to a 
preference scale (higher = better). We fit LOWESS smoothing \citep{cleveland1981lowess} with bandwidth $\text{frac}=0.4$ to the \{\ddp, preference\} pairs, a non-parametric local regression that estimates the conditional mean at each \ddp\ value without assuming a global functional form, suited to the unevenly distributed and potentially non-linear \ddp\ spectrum. We visualize both the absolute preference of each strategy and its difference relative to \APEseq, and report standard metric-based performance stratified by \ddp\ to show how an established evaluation frame yields new findings when organized along discourse dependency. Post-edit volume is reported as an auxiliary diagnostic with \ter. A qualitative analysis of cases where context-aware post-editing fails and where \ddp\ captures the underlying difficulty is provided in \Cref{appx:qualitative}.

\section{Results}
\label{sec:results}

\subsection{\ddp as a Predictor of Context Injection}

\paragraph{Strategy preference by human.}
We conduct relative ranking on two non-overlapping subsets of ${\approx}$500
segments, one per assumption. Because documents are selected by
inter-condition divergence rather than uniformly (\Cref{appx:humeval}), these
subsets over-represent the high-dependency region. 163 of the 498 \Aone\
segments (32.7\%) have $\ddp \geq 15$, against 212 of 849 (25.0\%) in the full
benchmark.

\Cref{fig:ddp_lowess} reveals consistent patterns across both assumptions. At low \ddp\ ($\lesssim$10), multiple APE strategies surpass \HPE, indicating that for segments requiring little discourse context, LLMs produce post-edits that annotators prefer over professional human revisions. From \ddp$\approx$10 onward, \HPE\ shows a monotonically increasing preference across the entire spectrum, while no APE strategy catches up.

Within APE strategies, \APEseq\ consistently outperforms \APEfull\ across nearly the full \ddp\ range (\Cref{fig:ddp_lowess}b). Providing the entire document does not help models exploit context more effectively than providing only the $k$ preceding segments, indicating that current LLMs do not extract useful signal from long-range document context even when it is available. \APEseg\ occasionally matches or outperforms \APEseq\ at low \ddp\ despite providing no context, which is consistent with the definition of low-\ddp\ segments as self-contained: injecting sequential context introduces noise rather than signal when no signal is needed.

In \Cref{fig:ddp_lowess}e, \APEseq\ is preferred over both \APErel\ and \APEexp\ across the \ddp\ spectrum. The advantage over \APEexp\ shows that providing context in its natural sentence form is more effective than abstracting it into structured declarative knowledge, and the near-identical trajectories of \APErel\ and \APEexp\ further suggest that what matters is not how segments are selected but the form in which context is delivered.

\begin{figure*}
    \centering
    \includegraphics[width=0.96\linewidth]{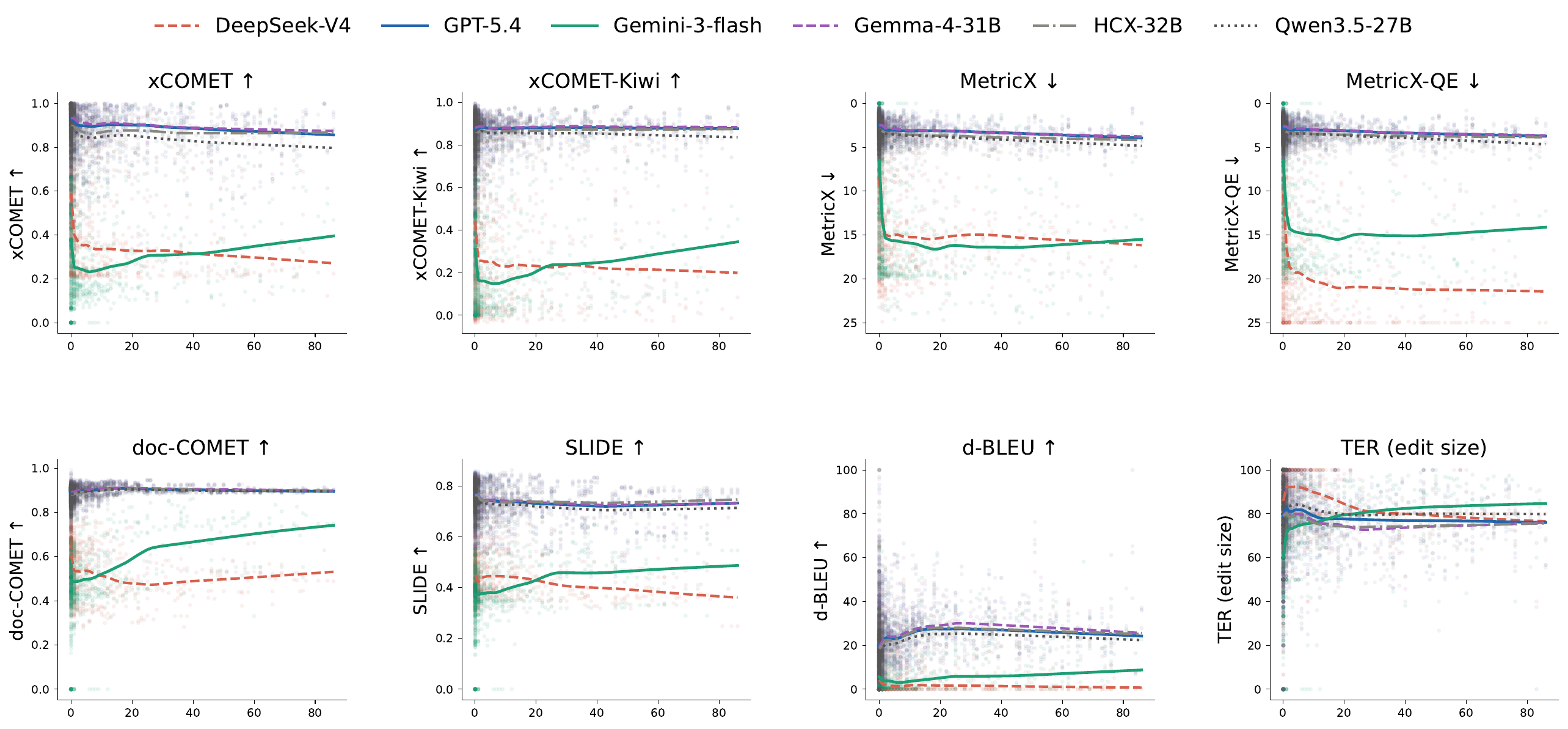}
    \caption{LOWESS-smoothed metric scores per model as a function of segment-level \ddp\ ($n = 229{,}230$ segment--model--strategy--metric observations). Models cluster into two groups across all metrics regardless of \ddp. \ter (bottom right) decreases as \ddp\ grows across most models, reflecting more conservative outputs in contextually demanding segments.
    }
    \label{fig:ddp_model}
\end{figure*}

\paragraph{Strategy preference by metrics.}
Across all automatic metrics (\Cref{fig:ddp_metrics} in \Cref{appx:results}),
strategy preferences remain largely insensitive to \ddp, in contrast to human
annotators whose preferences sharpen as discourse dependency grows
(\Cref{fig:ddp_lowess}c,f). Two patterns hold throughout. First, metric scores
show little directional change as \ddp\ increases, indicating that automatic
metrics do not register the growing contextual difficulty that human
evaluators respond to. Second, all APE strategies receive near-identical
scores across the \ddp\ spectrum, indicating that metrics cannot distinguish
strategies that humans clearly differentiate. QE metrics (\xcometkiwi,
\metricxqe) exhibit the flattest trajectories, lacking access to references
and missing even the surface-level differences that reference-based metrics
detect. \xcomet\ further ranks \APEfull\ above \APEseq\ across the entire
\ddp\ spectrum, reversing the human preference.

\paragraph{Statistical confirmation.}
The widening human preference for \HPE\ holds under significance testing.
Over the annotated segments the per-segment preference gap between \HPE\ and
\APEseq\ correlates positively with \ddp\ (Spearman $\rho = 0.28$,
$p = 1.2\times10^{-9}$, $n = 498$), so the effect is not confined to the tail.
Restricting to $\ddp \geq 15$, raters prefer \HPE\ over \APEseq\ by a mean of
$+0.35$ on the 1--4 scale (Wilcoxon signed-rank, one-sided
$p = 3.0\times10^{-5}$, bootstrap $95\%$ CI $[+0.18, +0.52]$, with \HPE\
ranked higher in $60\%$ of segments), whereas below $\ddp = 15$ the gap is
$+0.05$ and not significant. On the identical segments \xcomet\ cannot
separate the two (Wilcoxon $p = 0.40$, mean gap $-0.005$) and shows no trend
in \ddp\ (Spearman $\rho = 0.03$, $p = 0.50$). Automatic metrics should
therefore be interpreted with caution in document-level APE evaluation,
particularly at high \ddp, and discourse-stratified analyses provide a useful
complement to aggregate scores.

\subsection{\ddp as an Analytical Lens}

\paragraph{\ddp-level performance.}
Treating automatic metrics as a useful signal for system-level comparison, \ddp\ reveals patterns that aggregate scores obscure (\Cref{fig:ddp_model}). Models cluster into two groups regardless of \ddp: \mGPT, \mGemma, \mHCX, and \mQwen outperform the rest across the entire spectrum. Within the upper cluster, the four models produce nearly indistinguishable scores, suggesting that current metrics lack the resolution to rank closely matched systems. \mGemini is one exception, showing a mild upward trajectory on \textsc{COMET}-series as \ddp grows, while \mDeepSeek trends in the opposite direction. Separating segment- and document-level metrics reveals no qualitatively different patterns, except that \dbleu yields rankings consistent with neural metrics despite operating at the surface level.

\begin{table}[t]
\small
\centering
\resizebox{\linewidth}{!}{
\begin{tabular}{lrrrrrr}
\toprule
\textbf{Model} & \textbf{Overall} & \multicolumn{4}{c}{\textbf{By \ddp\ stratum (\%)}} \\
\cmidrule(lr){3-6}
 & \textbf{(\%)} & $=0$ & $1$--$4$ & $5$--$14$ & $15+$ \\
\midrule
\mDeepSeek    & 3.2  & 5.1  & 1.9  & 1.9  & 0.7 \\
\textsc{HCX}      & 7.2  & 10.1 & 6.4  & 4.5  & 2.8 \\
\mGemma    & 10.6 & 13.2 & 12.3 & 6.4  & 4.2 \\
\mGPT       & 11.5 & 13.8 & 16.0 & 6.4  & 2.8 \\
\mQwen    & 13.8 & 17.7 & 13.2 & 10.0 & 7.0 \\
\rowcolor{yellow!15} \mGemini & 26.9 & 44.2 & 16.9 & 15.5 & 4.9 \\
\bottomrule
\end{tabular}
}
\caption{Generation failure rate (\ter~$>$~100) per model and \ddp\ stratum. Failure rates consistently decrease as \ddp increases across all models.
}
\label{tab:hallucination}
\end{table}

\paragraph{Post-editing volume.}
\ter decreases as \ddp\ grows across most models and stabilizes beyond \ddp$\approx$20 (\Cref{fig:ddp_model}, bottom right), indicating that models produce more conservative outputs for contextually demanding segments. \mGemini exhibits the highest generation failure rate (26.9\%), concentrated at low \ddp\ (44.2\% at $\ddpop=0$) and dropping sharply at higher dependency. Since failures are excluded as generation errors and are concentrated at low \ddp, part of the decline in \ter\ reflects this differential filtering. We therefore read \ter\ as a diagnostic of edit volume rather than as a quality signal.

\section{Conclusion}
We introduced discourse dependency (\ddp), a metric-free, source-side measure of contextual difficulty in MT, and used it to revisit how context interacts with translation quality. 
Three implications follow. First, the gap is not a matter of input length. Models can ingest the full document but cannot selectively attend to the antecedents that resolve referential dependencies, while $k$-preceding injection often omits it entirely. Discourse-aware context selection, not context size, is the bottleneck. Second, metric blindness to \ddp\ is not incidental. Reference- and QE-based metrics are trained on parallel data dominated by sentence-level alignments, where inter-sentential phenomena such as coreference and entity tracking are rare and weakly supervised. Third, organizing evaluation by discourse dependency rather than topical domain admits a sharper accounting of progress, separating segments that current systems handle from those that genuinely demand discourse-level reasoning, before rather than after system outputs are produced. By making discourse dependency explicit in the source, \ddp\ offers a principled axis for evaluating what frontier MT systems actually understand about the documents they translate.

\section*{Limitations}
\label{sec:limit}

\paragraph{Single language pair.}
We focus on English-Korean as the primary testbed, a typologically distant pair where discourse phenomena such as pro-drop, honorifics, and named-entity rendering make cross-segment context consequential. A partial replication on English-Chinese (\Cref{appx:enzh}) shows broadly consistent metric trends, suggesting that the insensitivity to \ddp\ is not specific to Korean. Conclusive evidence requires human evaluation on additional pairs, which we leave for future work. How \ddp-stratified patterns generalize to other typologically distant pairs, to more closely related pairs, and to non-English source languages remains an important open question.

\paragraph{Anchor coverage.}
\ddp\ operationalizes contextual dependency through two anchor types: named-entity re-mentions and pronominal coreference. These cover a substantial portion of inter-segmental discourse dependency in English, but other phenomena such as lexical cohesion, tense and aspect consistency, ellipsis, and bridging anaphora fall outside the current definition. Our gender-number grouping also collapses singular gender-neutral \textit{they} (e.g., for non-binary referents) into the plural class, since surface form alone cannot disambiguate the two uses. The variation \ddp already exposes within and across domains suggests that even this minimal definition is informative; extending the anchor inventory and pronoun typology is a natural direction for future work.

\paragraph{Dependence on the NLP toolchain.}
Given a fixed anchor definition, \ddp\ relies on automatic NER, POS tagging, and dependency parsing. Pipeline errors are strongly one-directional, and a missed link shortens a chain and lowers \ddp, whereas inflation requires a spurious \emph{early} mention, which is far rarer. Our validation against gold coreference (\Cref{app:ontonotes}) confirms this in practice. \ddp\ under- or exactly estimates the gold distance in $99.2\%$ of segments, and the residual $0.8\%$ is traceable to NER false positives. The lower-bound reading is therefore an empirical near-guarantee rather than an absolute one, and it is tight enough that high-\ddp\ segments can be treated as certified to require long-range context. The ablation in \Cref{appx:ablation} further shows that architecturally distinct taggers yield strongly correlated values (Pearson $r=0.89$ for spaCy vs.\ Flair) and preserve the domain-level ordering.

\paragraph{Data contamination.}
WMT24++ was released in February 2025 and may overlap with the training data of recent LLMs. We mitigate this risk by adopting an APE setup. The models receive an initial machine translation and revise it rather than generating from the source alone, which reduces the benefit of memorized references. We cannot fully rule out residual contamination, particularly at low \ddp\ where surface memorization is most plausible. The main finding of this paper, that human preference for \HPE\ grows monotonically with \ddp, is least likely to be driven by contamination, since contamination would inflate model output quality rather than diminish it relative to human revisions.

\section*{Acknowledgment}
This work was supported by the G-LAMP Program of the National Research Foundation of Korea (NRF) grant funded by the Ministry of Education (No. RS-2025-25441317); the Ministry of Science and ICT (MSIT), and the National IT Industry Promotion Agency (NIPA) through the Advanced GPU Utilization Support Program (02-26-01-0499). 
The MSIT under the Convergence security core talent training business support program (IITP-2024 2024-RS-2024-00426853) supervised by the IITP(Institute of Information \& Communications Technology Planning \& Evaluation).
This work was supported by Korea Internet \& Security Agency(KISA) grant funded by the Korea government (PIPC) (No.RS-2026-25526342, Development of Technologies for Preventing Sensitive Information Inference and Risk Assessment in Foundation Model Operations)

\section*{Ethics Statement}
Our study involves expert human annotators. Participation was voluntary, and annotators were compensated at a fair market rate. All annotations were anonymized, and no personally identifiable information was collected.

Licenses of artifacts. All datasets, models, and software used in this study are publicly available for research purposes. The WMT24++ and WM25  test set is distributed under CC-BY 4.0; the automatic metrics are released under Apache 2.0. \mGPT is used via the OpenAI API under OpenAI's terms of service. Our benchmark and code will be released under CC-BY 4.0 and MIT, respectively.

Intended use. Our use of the WMT24++ and WM25 test set and automatic metrics is consistent with their intended use for MT evaluation research. The annotations and datasets we release are intended for research use only, consistent with the access conditions of the source data.

We acknowledge the use of AI assistants (\mClaude) for writing refinement and code review during paper preparation.

\bibliography{anthology-2,custom,arxiv/arxiv_bib}

@inproceedings{callison-burch-etal-2007-meta,title = "(Meta-) Evaluation of Machine Translation",author = "Callison-Burch, Chris and Fordyce, Cameron and Koehn, Philipp and Monz, Christof and Schroeder, Josh",editor = "Callison-Burch, Chris and Koehn, Philipp and Fordyce, Cameron Shaw and Monz, Christof",booktitle = "Proceedings of the Second Workshop on Statistical Machine Translation",month = jun,year = "2007",address = "Prague, Czech Republic",publisher = acl,url = anth # {W07-0718/},pages = "136--158"}

@inproceedings{hovy-etal-2006-ontonotes,title = "{O}nto{N}otes: The 90{\%} Solution",author = "Hovy, Eduard and Marcus, Mitchell and Palmer, Martha and Ramshaw, Lance and Weischedel, Ralph",editor = "Moore, Robert C. and Bilmes, Jeff and Chu-Carroll, Jennifer and Sanderson, Mark",booktitle = "Proceedings of the Human Language Technology Conference of the {NAACL}, Companion Volume: Short Papers",month = jun,year = "2006",address = "New York City, USA",publisher = acl,url = anth # {N06-2015/},pages = "57--60"}

@inproceedings{snover-etal-2006-study,title = "A Study of Translation Edit Rate with Targeted Human Annotation",author = "Snover, Matthew and Dorr, Bonnie and Schwartz, Rich and Micciulla, Linnea and Makhoul, John",booktitle = "Proceedings of the 7th Conference of the Association for Machine Translation in the Americas: Technical Papers",month = aug # " 8-12",year = "2006",address = "Cambridge, Massachusetts, USA",publisher = "Association for Machine Translation in the Americas",url = anth # {2006.amta-papers.25/},pages = "223--231"}

@article{grosz-etal-1995-centering,title = "{C}entering: A Framework for Modeling the Local Coherence of Discourse",author = "Grosz, Barbara J. and Joshi, Aravind K. and Weinstein, Scott",editor = "Hirschberg, Julia",journal = "Computational Linguistics",volume = "21",number = "2",year = "1995",address = "Cambridge, MA",publisher = "MIT Press",url = anth # {J95-2003/},pages = "203--225"}

@inproceedings{kocmi-etal-2024-findings,
    title = "Findings of the {WMT}24 General Machine Translation Shared Task: The {LLM} Era Is Here but {MT} Is Not Solved Yet",
    author = "Kocmi, Tom  and
      Avramidis, Eleftherios  and
      Bawden, Rachel  and
      Bojar, Ond{\v{r}}ej  and
      Dvorkovich, Anton  and
      Federmann, Christian  and
      Fishel, Mark  and
      Freitag, Markus  and
      Gowda, Thamme  and
      Grundkiewicz, Roman  and
      Haddow, Barry  and
      Karpinska, Marzena  and
      Koehn, Philipp  and
      Marie, Benjamin  and
      Monz, Christof  and
      Murray, Kenton  and
      Nagata, Masaaki  and
      Popel, Martin  and
      Popovi{\'c}, Maja  and
      Shmatova, Mariya  and
      Steingr{\'i}msson, Steinth{\'o}r  and
      Zouhar, Vil{\'e}m",
    editor = "Haddow, Barry  and
      Kocmi, Tom  and
      Koehn, Philipp  and
      Monz, Christof",
    booktitle = "Proceedings of the Ninth Conference on Machine Translation",
    month = nov,
    year = "2024",
    address = "Miami, Florida, USA",
    publisher = "Association for Computational Linguistics",
    url = "https://aclanthology.org/2024.wmt-1.1/",
    doi = "10.18653/v1/2024.wmt-1.1",
    pages = "1--46"
}

@inproceedings{kocmi-etal-2025-findings,
    title = "Findings of the {WMT}25 General Machine Translation Shared Task: Time to Stop Evaluating on Easy Test Sets",
    author = "Kocmi, Tom  and
      Artemova, Ekaterina  and
      Avramidis, Eleftherios  and
      Bawden, Rachel  and
      Bojar, Ond{\v{r}}ej  and
      Dranch, Konstantin  and
      Dvorkovich, Anton  and
      Dukanov, Sergey  and
      Fishel, Mark  and
      Freitag, Markus  and
      Gowda, Thamme  and
      Grundkiewicz, Roman  and
      Haddow, Barry  and
      Karpinska, Marzena  and
      Koehn, Philipp  and
      Lakougna, Howard  and
      Lundin, Jessica  and
      Monz, Christof  and
      Murray, Kenton  and
      Nagata, Masaaki  and
      Perrella, Stefano  and
      Proietti, Lorenzo  and
      Popel, Martin  and
      Popovi{\'c}, Maja  and
      Riley, Parker  and
      Shmatova, Mariya  and
      Steingr{\'i}msson, Steinth{\'o}r  and
      Yankovskaya, Lisa  and
      Zouhar, Vil{\'e}m",
    editor = "Haddow, Barry  and
      Kocmi, Tom  and
      Koehn, Philipp  and
      Monz, Christof",
    booktitle = "Proceedings of the Tenth Conference on Machine Translation",
    month = nov,
    year = "2025",
    address = "Suzhou, China",
    publisher = "Association for Computational Linguistics",
    url = "https://aclanthology.org/2025.wmt-1.22/",
    doi = "10.18653/v1/2025.wmt-1.22",
    pages = "355--413",
    ISBN = "979-8-89176-341-8"
}

@inproceedings{proietti-etal-2025-estimating,
    title = "Estimating Machine Translation Difficulty",
    author = "Proietti, Lorenzo  and
      Perrella, Stefano  and
      Zouhar, Vil{\'e}m  and
      Navigli, Roberto  and
      Kocmi, Tom",
    editor = "Christodoulopoulos, Christos  and
      Chakraborty, Tanmoy  and
      Rose, Carolyn  and
      Peng, Violet",
    booktitle = "Findings of the Association for Computational Linguistics: EMNLP 2025",
    month = nov,
    year = "2025",
    address = "Suzhou, China",
    publisher = "Association for Computational Linguistics",
    url = "https://aclanthology.org/2025.findings-emnlp.1317/",
    doi = "10.18653/v1/2025.findings-emnlp.1317",
    pages = "24261--24285",
    ISBN = "979-8-89176-335-7"
}

@inproceedings{mathur-etal-2020-tangled,
    title = "Tangled up in {BLEU}: Reevaluating the Evaluation of Automatic Machine Translation Evaluation Metrics",
    author = "Mathur, Nitika  and
      Baldwin, Timothy  and
      Cohn, Trevor",
    editor = "Jurafsky, Dan  and
      Chai, Joyce  and
      Schluter, Natalie  and
      Tetreault, Joel",
    booktitle = "Proceedings of the 58th Annual Meeting of the Association for Computational Linguistics",
    month = jul,
    year = "2020",
    address = "Online",
    publisher = "Association for Computational Linguistics",
    url = "https://aclanthology.org/2020.acl-main.448/",
    doi = "10.18653/v1/2020.acl-main.448",
    pages = "4984--4997"
}

@inproceedings{freitag-etal-2022-results,
    title = "Results of {WMT}22 Metrics Shared Task: Stop Using {BLEU} {--} Neural Metrics Are Better and More Robust",
    author = "Freitag, Markus  and
      Rei, Ricardo  and
      Mathur, Nitika  and
      Lo, Chi-kiu  and
      Stewart, Craig  and
      Avramidis, Eleftherios  and
      Kocmi, Tom  and
      Foster, George  and
      Lavie, Alon  and
      Martins, Andr{\'e} F. T.",
    editor = {Koehn, Philipp  and
      Barrault, Lo{\"i}c  and
      Bojar, Ond{\v{r}}ej  and
      Bougares, Fethi  and
      Chatterjee, Rajen  and
      Costa-juss{\`a}, Marta R.  and
      Federmann, Christian  and
      Fishel, Mark  and
      Fraser, Alexander  and
      Freitag, Markus  and
      Graham, Yvette  and
      Grundkiewicz, Roman  and
      Guzman, Paco  and
      Haddow, Barry  and
      Huck, Matthias  and
      Jimeno Yepes, Antonio  and
      Kocmi, Tom  and
      Martins, Andr{\'e}  and
      Morishita, Makoto  and
      Monz, Christof  and
      Nagata, Masaaki  and
      Nakazawa, Toshiaki  and
      Negri, Matteo  and
      N{\'e}v{\'e}ol, Aur{\'e}lie  and
      Neves, Mariana  and
      Popel, Martin  and
      Turchi, Marco  and
      Zampieri, Marcos},
    booktitle = "Proceedings of the Seventh Conference on Machine Translation (WMT)",
    month = dec,
    year = "2022",
    address = "Abu Dhabi, United Arab Emirates (Hybrid)",
    publisher = "Association for Computational Linguistics",
    url = "https://aclanthology.org/2022.wmt-1.2/",
    doi = "10.18653/v1/2022.wmt-1.2",
    pages = "46--68"
}

@inproceedings{lavie-etal-2025-findings,
    title = "Findings of the {WMT}25 Shared Task on Automated Translation Evaluation Systems: Linguistic Diversity is Challenging and References Still Help",
    author = "Lavie, Alon  and
      Hanneman, Greg  and
      Agrawal, Sweta  and
      Kanojia, Diptesh  and
      Lo, Chi-Kiu  and
      Zouhar, Vil{\'e}m  and
      Blain, Frederic  and
      Zerva, Chrysoula  and
      Avramidis, Eleftherios  and
      Deoghare, Sourabh  and
      Sindhujan, Archchana  and
      Wang, Jiayi  and
      Adelani, David Ifeoluwa  and
      Thompson, Brian  and
      Kocmi, Tom  and
      Freitag, Markus  and
      Deutsch, Daniel",
    editor = "Haddow, Barry  and
      Kocmi, Tom  and
      Koehn, Philipp  and
      Monz, Christof",
    booktitle = "Proceedings of the Tenth Conference on Machine Translation",
    month = nov,
    year = "2025",
    address = "Suzhou, China",
    publisher = "Association for Computational Linguistics",
    url = "https://aclanthology.org/2025.wmt-1.24/",
    doi = "10.18653/v1/2025.wmt-1.24",
    pages = "436--483",
    ISBN = "979-8-89176-341-8"
}

@inproceedings{voita-etal-2019-good,
    title = "When a Good Translation is Wrong in Context: Context-Aware Machine Translation Improves on Deixis, Ellipsis, and Lexical Cohesion",
    author = "Voita, Elena  and
      Sennrich, Rico  and
      Titov, Ivan",
    editor = "Korhonen, Anna  and
      Traum, David  and
      M{\`a}rquez, Llu{\'i}s",
    booktitle = "Proceedings of the 57th Annual Meeting of the Association for Computational Linguistics",
    month = jul,
    year = "2019",
    address = "Florence, Italy",
    publisher = "Association for Computational Linguistics",
    url = "https://aclanthology.org/P19-1116/",
    doi = "10.18653/v1/P19-1116",
    pages = "1198--1212"
}

@inproceedings{bawden-etal-2018-evaluating,
    title = "Evaluating Discourse Phenomena in Neural Machine Translation",
    author = "Bawden, Rachel  and
      Sennrich, Rico  and
      Birch, Alexandra  and
      Haddow, Barry",
    editor = "Walker, Marilyn  and
      Ji, Heng  and
      Stent, Amanda",
    booktitle = "Proceedings of the 2018 Conference of the North {A}merican Chapter of the Association for Computational Linguistics: Human Language Technologies, Volume 1 (Long Papers)",
    month = jun,
    year = "2018",
    address = "New Orleans, Louisiana",
    publisher = "Association for Computational Linguistics",
    url = "https://aclanthology.org/N18-1118/",
    doi = "10.18653/v1/N18-1118",
    pages = "1304--1313"
}

@inproceedings{fernandes-etal-2021-measuring,
    title = "Measuring and Increasing Context Usage in Context-Aware Machine Translation",
    author = "Fernandes, Patrick  and
      Yin, Kayo  and
      Neubig, Graham  and
      Martins, Andr{\'e} F. T.",
    editor = "Zong, Chengqing  and
      Xia, Fei  and
      Li, Wenjie  and
      Navigli, Roberto",
    booktitle = "Proceedings of the 59th Annual Meeting of the Association for Computational Linguistics and the 11th International Joint Conference on Natural Language Processing (Volume 1: Long Papers)",
    month = aug,
    year = "2021",
    address = "Online",
    publisher = "Association for Computational Linguistics",
    url = "https://aclanthology.org/2021.acl-long.505/",
    doi = "10.18653/v1/2021.acl-long.505",
    pages = "6467--6478"
}

@inproceedings{fernandes-etal-2023-translation,
    title = "When Does Translation Require Context? A Data-driven, Multilingual Exploration",
    author = "Fernandes, Patrick  and
      Yin, Kayo  and
      Liu, Emmy  and
      Martins, Andr{\'e}  and
      Neubig, Graham",
    editor = "Rogers, Anna  and
      Boyd-Graber, Jordan  and
      Okazaki, Naoaki",
    booktitle = "Proceedings of the 61st Annual Meeting of the Association for Computational Linguistics (Volume 1: Long Papers)",
    month = jul,
    year = "2023",
    address = "Toronto, Canada",
    publisher = "Association for Computational Linguistics",
    url = "https://aclanthology.org/2023.acl-long.36/",
    doi = "10.18653/v1/2023.acl-long.36",
    pages = "606--626"
}

@inproceedings{wang-etal-2023-document-level,
    title = "Document-Level Machine Translation with Large Language Models",
    author = "Wang, Longyue  and
      Lyu, Chenyang  and
      Ji, Tianbo  and
      Zhang, Zhirui  and
      Yu, Dian  and
      Shi, Shuming  and
      Tu, Zhaopeng",
    editor = "Bouamor, Houda  and
      Pino, Juan  and
      Bali, Kalika",
    booktitle = "Proceedings of the 2023 Conference on Empirical Methods in Natural Language Processing",
    month = dec,
    year = "2023",
    address = "Singapore",
    publisher = "Association for Computational Linguistics",
    url = "https://aclanthology.org/2023.emnlp-main.1036/",
    doi = "10.18653/v1/2023.emnlp-main.1036",
    pages = "16646--16661"
}

@inproceedings{park-pado-2024-multi,
    title = "Multi-Dimensional Machine Translation Evaluation: Model Evaluation and Resource for {K}orean",
    author = "Park, Dojun  and
      Pad{\'o}, Sebastian",
    editor = "Calzolari, Nicoletta  and
      Kan, Min-Yen  and
      Hoste, Veronique  and
      Lenci, Alessandro  and
      Sakti, Sakriani  and
      Xue, Nianwen",
    booktitle = "Proceedings of the 2024 Joint International Conference on Computational Linguistics, Language Resources and Evaluation (LREC-COLING 2024)",
    month = may,
    year = "2024",
    address = "Torino, Italia",
    publisher = "ELRA and ICCL",
    url = "https://aclanthology.org/2024.lrec-main.1024/",
    pages = "11723--11744"
}

@inproceedings{choi-etal-2018-automatic,
    title = "Automatic Evaluation of {E}nglish-to-{K}orean and {K}orean-to-{E}nglish Neural Machine Translation Systems by Linguistic Test Points",
    author = "Choi, Sung-Kwon  and
      Choi, Gyu-Hyeun  and
      Kim, Youngkil",
    editor = "Politzer-Ahles, Stephen  and
      Hsu, Yu-Yin  and
      Huang, Chu-Ren  and
      Yao, Yao",
    booktitle = "Proceedings of the 32nd Pacific Asia Conference on Language, Information and Computation",
    month = "1–3 " # dec,
    year = "2018",
    address = "Hong Kong",
    publisher = "Association for Computational Linguistics",
    url = "https://aclanthology.org/Y18-1013/"
}

@inproceedings{lee-etal-2025-testset,
    title = "A Testset for Context-Aware {LLM} Translation in {K}orean-to-{E}nglish Discourse Level Translation",
    author = "Lee, Minjae  and
      Noh, Youngbin  and
      Lee, Seung Jin",
    editor = "Rambow, Owen  and
      Wanner, Leo  and
      Apidianaki, Marianna  and
      Al-Khalifa, Hend  and
      Eugenio, Barbara Di  and
      Schockaert, Steven",
    booktitle = "Proceedings of the 31st International Conference on Computational Linguistics",
    month = jan,
    year = "2025",
    address = "Abu Dhabi, UAE",
    publisher = "Association for Computational Linguistics",
    url = "https://aclanthology.org/2025.coling-main.110/",
    pages = "1632--1646"
}

@inproceedings{barrault-etal-2019-findings,
    title = "Findings of the 2019 Conference on Machine Translation ({WMT}19)",
    author = {Barrault, Lo{\"i}c  and
      Bojar, Ond{\v{r}}ej  and
      Costa-juss{\`a}, Marta R.  and
      Federmann, Christian  and
      Fishel, Mark  and
      Graham, Yvette  and
      Haddow, Barry  and
      Huck, Matthias  and
      Koehn, Philipp  and
      Malmasi, Shervin  and
      Monz, Christof  and
      M{\"u}ller, Mathias  and
      Pal, Santanu  and
      Post, Matt  and
      Zampieri, Marcos},
    editor = "Bojar, Ond{\v{r}}ej  and
      Chatterjee, Rajen  and
      Federmann, Christian  and
      Fishel, Mark  and
      Graham, Yvette  and
      Haddow, Barry  and
      Huck, Matthias  and
      Yepes, Antonio Jimeno  and
      Koehn, Philipp  and
      Martins, Andr{\'e}  and
      Monz, Christof  and
      Negri, Matteo  and
      N{\'e}v{\'e}ol, Aur{\'e}lie  and
      Neves, Mariana  and
      Post, Matt  and
      Turchi, Marco  and
      Verspoor, Karin",
    booktitle = "Proceedings of the Fourth Conference on Machine Translation (Volume 2: Shared Task Papers, Day 1)",
    month = aug,
    year = "2019",
    address = "Florence, Italy",
    publisher = "Association for Computational Linguistics",
    url = "https://aclanthology.org/W19-5301/",
    doi = "10.18653/v1/W19-5301",
    pages = "1--61"
}

@inproceedings{akhbardeh-etal-2021-findings,
    title = "Findings of the 2021 Conference on Machine Translation ({WMT}21)",
    author = "Akhbardeh, Farhad  and
      Arkhangorodsky, Arkady  and
      Biesialska, Magdalena  and
      Bojar, Ond{\v{r}}ej  and
      Chatterjee, Rajen  and
      Chaudhary, Vishrav  and
      Costa-jussa, Marta R.  and
      Espa{\~n}a-Bonet, Cristina  and
      Fan, Angela  and
      Federmann, Christian  and
      Freitag, Markus  and
      Graham, Yvette  and
      Grundkiewicz, Roman  and
      Haddow, Barry  and
      Harter, Leonie  and
      Heafield, Kenneth  and
      Homan, Christopher M.  and
      Huck, Matthias  and
      Amponsah-Kaakyire, Kwabena  and
      Kasai, Jungo  and
      Khashabi, Daniel  and
      Knight, Kevin  and
      Kocmi, Tom  and
      Koehn, Philipp  and
      Lourie, Nicholas  and
      Monz, Christof  and
      Morishita, Makoto  and
      Nagata, Masaaki  and
      Nagesh, Ajay  and
      Nakazawa, Toshiaki  and
      Negri, Matteo  and
      Pal, Santanu  and
      Tapo, Allahsera Auguste  and
      Turchi, Marco  and
      Vydrin, Valentin  and
      Zampieri, Marcos",
    editor = "Barrault, Loic  and
      Bojar, Ondrej  and
      Bougares, Fethi  and
      Chatterjee, Rajen  and
      Costa-jussa, Marta R.  and
      Federmann, Christian  and
      Fishel, Mark  and
      Fraser, Alexander  and
      Freitag, Markus  and
      Graham, Yvette  and
      Grundkiewicz, Roman  and
      Guzman, Paco  and
      Haddow, Barry  and
      Huck, Matthias  and
      Yepes, Antonio Jimeno  and
      Koehn, Philipp  and
      Kocmi, Tom  and
      Martins, Andre  and
      Morishita, Makoto  and
      Monz, Christof",
    booktitle = "Proceedings of the Sixth Conference on Machine Translation",
    month = nov,
    year = "2021",
    address = "Online",
    publisher = "Association for Computational Linguistics",
    url = "https://aclanthology.org/2021.wmt-1.1/",
    pages = "1--88"
}

@inproceedings{kocmi-etal-2022-findings,
    title = "Findings of the 2022 Conference on Machine Translation ({WMT}22)",
    author = "Kocmi, Tom  and
      Bawden, Rachel  and
      Bojar, Ond{\v{r}}ej  and
      Dvorkovich, Anton  and
      Federmann, Christian  and
      Fishel, Mark  and
      Gowda, Thamme  and
      Graham, Yvette  and
      Grundkiewicz, Roman  and
      Haddow, Barry  and
      Knowles, Rebecca  and
      Koehn, Philipp  and
      Monz, Christof  and
      Morishita, Makoto  and
      Nagata, Masaaki  and
      Nakazawa, Toshiaki  and
      Nov{\'a}k, Michal  and
      Popel, Martin  and
      Popovi{\'c}, Maja",
    editor = {Koehn, Philipp  and
      Barrault, Lo{\"i}c  and
      Bojar, Ond{\v{r}}ej  and
      Bougares, Fethi  and
      Chatterjee, Rajen  and
      Costa-juss{\`a}, Marta R.  and
      Federmann, Christian  and
      Fishel, Mark  and
      Fraser, Alexander  and
      Freitag, Markus  and
      Graham, Yvette  and
      Grundkiewicz, Roman  and
      Guzman, Paco  and
      Haddow, Barry  and
      Huck, Matthias  and
      Jimeno Yepes, Antonio  and
      Kocmi, Tom  and
      Martins, Andr{\'e}  and
      Morishita, Makoto  and
      Monz, Christof  and
      Nagata, Masaaki  and
      Nakazawa, Toshiaki  and
      Negri, Matteo  and
      N{\'e}v{\'e}ol, Aur{\'e}lie  and
      Neves, Mariana  and
      Popel, Martin  and
      Turchi, Marco  and
      Zampieri, Marcos},
    booktitle = "Proceedings of the Seventh Conference on Machine Translation (WMT)",
    month = dec,
    year = "2022",
    address = "Abu Dhabi, United Arab Emirates (Hybrid)",
    publisher = "Association for Computational Linguistics",
    url = "https://aclanthology.org/2022.wmt-1.1/",
    doi = "10.18653/v1/2022.wmt-1.1",
    pages = "1--45"
}

@inproceedings{muller-etal-2018-large,
    title = "A Large-Scale Test Set for the Evaluation of Context-Aware Pronoun Translation in Neural Machine Translation",
    author = {M{\"u}ller, Mathias  and
      Rios, Annette  and
      Voita, Elena  and
      Sennrich, Rico},
    editor = "Bojar, Ond{\v{r}}ej  and
      Chatterjee, Rajen  and
      Federmann, Christian  and
      Fishel, Mark  and
      Graham, Yvette  and
      Haddow, Barry  and
      Huck, Matthias  and
      Yepes, Antonio Jimeno  and
      Koehn, Philipp  and
      Monz, Christof  and
      Negri, Matteo  and
      N{\'e}v{\'e}ol, Aur{\'e}lie  and
      Neves, Mariana  and
      Post, Matt  and
      Specia, Lucia  and
      Turchi, Marco  and
      Verspoor, Karin",
    booktitle = "Proceedings of the Third Conference on Machine Translation: Research Papers",
    month = oct,
    year = "2018",
    address = "Brussels, Belgium",
    publisher = "Association for Computational Linguistics",
    url = "https://aclanthology.org/W18-6307/",
    doi = "10.18653/v1/W18-6307",
    pages = "61--72"
}

@inproceedings{kocmi-etal-2023-findings,
    title = "Findings of the 2023 Conference on Machine Translation ({WMT}23): {LLM}s Are Here but Not Quite There Yet",
    author = "Kocmi, Tom  and
      Avramidis, Eleftherios  and
      Bawden, Rachel  and
      Bojar, Ond{\v{r}}ej  and
      Dvorkovich, Anton  and
      Federmann, Christian  and
      Fishel, Mark  and
      Freitag, Markus  and
      Gowda, Thamme  and
      Grundkiewicz, Roman  and
      Haddow, Barry  and
      Koehn, Philipp  and
      Marie, Benjamin  and
      Monz, Christof  and
      Morishita, Makoto  and
      Murray, Kenton  and
      Nagata, Masaaki  and
      Nakazawa, Toshiaki  and
      Popel, Martin  and
      Popovi{\'c}, Maja  and
      Shmatova, Mariya  and
      Suzuki, Jun",
    editor = "Koehn, Philipp  and
      Haddow, Barry  and
      Kocmi, Tom  and
      Monz, Christof",
    booktitle = "Proceedings of the Eighth Conference on Machine Translation",
    month = dec,
    year = "2023",
    address = "Singapore",
    publisher = "Association for Computational Linguistics",
    url = "https://aclanthology.org/2023.wmt-1.1/",
    doi = "10.18653/v1/2023.wmt-1.1",
    pages = "1--42"
}

@inproceedings{manakhimova-etal-2024-investigating,
    title = "Investigating the Linguistic Performance of Large Language Models in Machine Translation",
    author = {Manakhimova, Shushen  and
      Macketanz, Vivien  and
      Avramidis, Eleftherios  and
      Lapshinova-Koltunski, Ekaterina  and
      Bagdasarov, Sergei  and
      M{\"o}ller, Sebastian},
    editor = "Haddow, Barry  and
      Kocmi, Tom  and
      Koehn, Philipp  and
      Monz, Christof",
    booktitle = "Proceedings of the Ninth Conference on Machine Translation",
    month = nov,
    year = "2024",
    address = "Miami, Florida, USA",
    publisher = "Association for Computational Linguistics",
    url = "https://aclanthology.org/2024.wmt-1.28/",
    doi = "10.18653/v1/2024.wmt-1.28",
    pages = "355--371"
}

@inproceedings{bhattacharjee-etal-2024-domain-dynamics,
    title = "Domain Dynamics: Evaluating Large Language Models in {E}nglish-{H}indi Translation",
    author = "Bhattacharjee, Soham  and
      Gain, Baban  and
      Ekbal, Asif",
    editor = "Lalitha Devi, Sobha  and
      Arora, Karunesh",
    booktitle = "Proceedings of the 21st International Conference on Natural Language Processing (ICON)",
    month = dec,
    year = "2024",
    address = "AU-KBC Research Centre, Chennai, India",
    publisher = "NLP Association of India (NLPAI)",
    url = "https://aclanthology.org/2024.icon-1.19/",
    pages = "169--177"
}

@inproceedings{wicks-post-2023-identifying,
    title = "Identifying Context-Dependent Translations for Evaluation Set Production",
    author = "Wicks, Rachel  and
      Post, Matt",
    editor = "Koehn, Philipp  and
      Haddow, Barry  and
      Kocmi, Tom  and
      Monz, Christof",
    booktitle = "Proceedings of the Eighth Conference on Machine Translation",
    month = dec,
    year = "2023",
    address = "Singapore",
    publisher = "Association for Computational Linguistics",
    url = "https://aclanthology.org/2023.wmt-1.42/",
    doi = "10.18653/v1/2023.wmt-1.42",
    pages = "452--467"
}

@inproceedings{mohammed-niculae-2024-measuring,
    title = "On Measuring Context Utilization in Document-Level {MT} Systems",
    author = "Mohammed, Wafaa  and
      Niculae, Vlad",
    editor = "Graham, Yvette  and
      Purver, Matthew",
    booktitle = "Findings of the Association for Computational Linguistics: EACL 2024",
    month = mar,
    year = "2024",
    address = "St. Julian{'}s, Malta",
    publisher = "Association for Computational Linguistics",
    url = "https://aclanthology.org/2024.findings-eacl.113/",
    doi = "10.18653/v1/2024.findings-eacl.113",
    pages = "1633--1643"
}

@inproceedings{tiedemann-scherrer-2017-neural,
    title = "Neural Machine Translation with Extended Context",
    author = {Tiedemann, J{\"o}rg  and
      Scherrer, Yves},
    editor = {Webber, Bonnie  and
      Popescu-Belis, Andrei  and
      Tiedemann, J{\"o}rg},
    booktitle = "Proceedings of the Third Workshop on Discourse in Machine Translation",
    month = sep,
    year = "2017",
    address = "Copenhagen, Denmark",
    publisher = "Association for Computational Linguistics",
    url = "https://aclanthology.org/W17-4811/",
    doi = "10.18653/v1/W17-4811",
    pages = "82--92"
}

@inproceedings{junczys-dowmunt-2019-microsoft,
    title = "{M}icrosoft Translator at {WMT} 2019: Towards Large-Scale Document-Level Neural Machine Translation",
    author = "Junczys-Dowmunt, Marcin",
    editor = "Bojar, Ond{\v{r}}ej  and
      Chatterjee, Rajen  and
      Federmann, Christian  and
      Fishel, Mark  and
      Graham, Yvette  and
      Haddow, Barry  and
      Huck, Matthias  and
      Yepes, Antonio Jimeno  and
      Koehn, Philipp  and
      Martins, Andr{\'e}  and
      Monz, Christof  and
      Negri, Matteo  and
      N{\'e}v{\'e}ol, Aur{\'e}lie  and
      Neves, Mariana  and
      Post, Matt  and
      Turchi, Marco  and
      Verspoor, Karin",
    booktitle = "Proceedings of the Fourth Conference on Machine Translation (Volume 2: Shared Task Papers, Day 1)",
    month = aug,
    year = "2019",
    address = "Florence, Italy",
    publisher = "Association for Computational Linguistics",
    url = "https://aclanthology.org/W19-5321/",
    doi = "10.18653/v1/W19-5321",
    pages = "225--233"
}

@inproceedings{sun-etal-2022-rethinking,
    title = "Rethinking Document-level Neural Machine Translation",
    author = "Sun, Zewei  and
      Wang, Mingxuan  and
      Zhou, Hao  and
      Zhao, Chengqi  and
      Huang, Shujian  and
      Chen, Jiajun  and
      Li, Lei",
    editor = "Muresan, Smaranda  and
      Nakov, Preslav  and
      Villavicencio, Aline",
    booktitle = "Findings of the Association for Computational Linguistics: ACL 2022",
    month = may,
    year = "2022",
    address = "Dublin, Ireland",
    publisher = "Association for Computational Linguistics",
    url = "https://aclanthology.org/2022.findings-acl.279/",
    doi = "10.18653/v1/2022.findings-acl.279",
    pages = "3537--3548"
}

@inproceedings{karpinska-iyyer-2023-large,
    title = "Large Language Models Effectively Leverage Document-level Context for Literary Translation, but Critical Errors Persist",
    author = "Karpinska, Marzena  and
      Iyyer, Mohit",
    editor = "Koehn, Philipp  and
      Haddow, Barry  and
      Kocmi, Tom  and
      Monz, Christof",
    booktitle = "Proceedings of the Eighth Conference on Machine Translation",
    month = dec,
    year = "2023",
    address = "Singapore",
    publisher = "Association for Computational Linguistics",
    url = "https://aclanthology.org/2023.wmt-1.41/",
    doi = "10.18653/v1/2023.wmt-1.41",
    pages = "419--451"
}

@inproceedings{akbik-etal-2019-flair,
    title = "{FLAIR}: An Easy-to-Use Framework for State-of-the-Art {NLP}",
    author = "Akbik, Alan  and
      Bergmann, Tanja  and
      Blythe, Duncan  and
      Rasul, Kashif  and
      Schweter, Stefan  and
      Vollgraf, Roland",
    editor = "Ammar, Waleed  and
      Louis, Annie  and
      Mostafazadeh, Nasrin",
    booktitle = "Proceedings of the 2019 Conference of the North {A}merican Chapter of the Association for Computational Linguistics (Demonstrations)",
    month = jun,
    year = "2019",
    address = "Minneapolis, Minnesota",
    publisher = "Association for Computational Linguistics",
    url = "https://aclanthology.org/N19-4010/",
    doi = "10.18653/v1/N19-4010",
    pages = "54--59"
}

@inproceedings{deutsch-etal-2025-wmt24,
    title = "{WMT}24++: Expanding the Language Coverage of {WMT}24 to 55 Languages {\&} Dialects",
    author = "Deutsch, Daniel  and
      Briakou, Eleftheria  and
      Caswell, Isaac Rayburn  and
      Finkelstein, Mara  and
      Galor, Rebecca  and
      Juraska, Juraj  and
      Kovacs, Geza  and
      Lui, Alison  and
      Rei, Ricardo  and
      Riesa, Jason  and
      Rijhwani, Shruti  and
      Riley, Parker  and
      Salesky, Elizabeth  and
      Trabelsi, Firas  and
      Winkler, Stephanie  and
      Zhang, Biao  and
      Freitag, Markus",
    editor = "Che, Wanxiang  and
      Nabende, Joyce  and
      Shutova, Ekaterina  and
      Pilehvar, Mohammad Taher",
    booktitle = "Findings of the Association for Computational Linguistics: ACL 2025",
    month = jul,
    year = "2025",
    address = "Vienna, Austria",
    publisher = "Association for Computational Linguistics",
    url = "https://aclanthology.org/2025.findings-acl.634/",
    doi = "10.18653/v1/2025.findings-acl.634",
    pages = "12257--12284",
    ISBN = "979-8-89176-256-5"
}

@inproceedings{rei-etal-2023-scaling,
    title = "Scaling up {C}omet{K}iwi: Unbabel-{IST} 2023 Submission for the Quality Estimation Shared Task",
    author = "Rei, Ricardo  and
      Guerreiro, Nuno M.  and
      Pombal, Jos{\'e}  and
      van Stigt, Daan  and
      Treviso, Marcos  and
      Coheur, Luisa  and
      C. de Souza, Jos{\'e} G.  and
      Martins, Andr{\'e} F. T.",
    editor = "Koehn, Philipp  and
      Haddow, Barry  and
      Kocmi, Tom  and
      Monz, Christof",
    booktitle = "Proceedings of the Eighth Conference on Machine Translation",
    month = dec,
    year = "2023",
    address = "Singapore",
    publisher = "Association for Computational Linguistics",
    url = "https://aclanthology.org/2023.wmt-1.73/",
    doi = "10.18653/v1/2023.wmt-1.73",
    pages = "841--848"
}

@inproceedings{juraska-etal-2024-metricx,
    title = "{M}etric{X}-24: The {G}oogle Submission to the {WMT} 2024 Metrics Shared Task",
    author = "Juraska, Juraj  and
      Deutsch, Daniel  and
      Finkelstein, Mara  and
      Freitag, Markus",
    editor = "Haddow, Barry  and
      Kocmi, Tom  and
      Koehn, Philipp  and
      Monz, Christof",
    booktitle = "Proceedings of the Ninth Conference on Machine Translation",
    month = nov,
    year = "2024",
    address = "Miami, Florida, USA",
    publisher = "Association for Computational Linguistics",
    url = "https://aclanthology.org/2024.wmt-1.35/",
    doi = "10.18653/v1/2024.wmt-1.35",
    pages = "492--504"
}

@inproceedings{vernikos-etal-2022-embarrassingly,
    title = "Embarrassingly Easy Document-Level {MT} Metrics: How to Convert Any Pretrained Metric into a Document-Level Metric",
    author = "Vernikos, Giorgos  and
      Thompson, Brian  and
      Mathur, Prashant  and
      Federico, Marcello",
    editor = {Koehn, Philipp  and
      Barrault, Lo{\"i}c  and
      Bojar, Ond{\v{r}}ej  and
      Bougares, Fethi  and
      Chatterjee, Rajen  and
      Costa-juss{\`a}, Marta R.  and
      Federmann, Christian  and
      Fishel, Mark  and
      Fraser, Alexander  and
      Freitag, Markus  and
      Graham, Yvette  and
      Grundkiewicz, Roman  and
      Guzman, Paco  and
      Haddow, Barry  and
      Huck, Matthias  and
      Jimeno Yepes, Antonio  and
      Kocmi, Tom  and
      Martins, Andr{\'e}  and
      Morishita, Makoto  and
      Monz, Christof  and
      Nagata, Masaaki  and
      Nakazawa, Toshiaki  and
      Negri, Matteo  and
      N{\'e}v{\'e}ol, Aur{\'e}lie  and
      Neves, Mariana  and
      Popel, Martin  and
      Turchi, Marco  and
      Zampieri, Marcos},
    booktitle = "Proceedings of the Seventh Conference on Machine Translation (WMT)",
    month = dec,
    year = "2022",
    address = "Abu Dhabi, United Arab Emirates (Hybrid)",
    publisher = "Association for Computational Linguistics",
    url = "https://aclanthology.org/2022.wmt-1.6/",
    doi = "10.18653/v1/2022.wmt-1.6",
    pages = "118--128"
}

@inproceedings{raunak-etal-2023-evaluating,
    title = "Evaluating Metrics for Document-context Evaluation in Machine Translation",
    author = "Raunak, Vikas  and
      Kocmi, Tom  and
      Post, Matt",
    editor = "Koehn, Philipp  and
      Haddow, Barry  and
      Kocmi, Tom  and
      Monz, Christof",
    booktitle = "Proceedings of the Eighth Conference on Machine Translation",
    month = dec,
    year = "2023",
    address = "Singapore",
    publisher = "Association for Computational Linguistics",
    url = "https://aclanthology.org/2023.wmt-1.68/",
    doi = "10.18653/v1/2023.wmt-1.68",
    pages = "812--814"
}

@inproceedings{feng-etal-2022-language,
    title = "Language-agnostic {BERT} Sentence Embedding",
    author = "Feng, Fangxiaoyu  and
      Yang, Yinfei  and
      Cer, Daniel  and
      Arivazhagan, Naveen  and
      Wang, Wei",
    editor = "Muresan, Smaranda  and
      Nakov, Preslav  and
      Villavicencio, Aline",
    booktitle = "Proceedings of the 60th Annual Meeting of the Association for Computational Linguistics (Volume 1: Long Papers)",
    month = may,
    year = "2022",
    address = "Dublin, Ireland",
    publisher = "Association for Computational Linguistics",
    url = "https://aclanthology.org/2022.acl-long.62/",
    doi = "10.18653/v1/2022.acl-long.62",
    pages = "878--891"
}

@inproceedings{miculicich-etal-2018-document,
    title = "Document-Level Neural Machine Translation with Hierarchical Attention Networks",
    author = "Miculicich, Lesly  and
      Ram, Dhananjay  and
      Pappas, Nikolaos  and
      Henderson, James",
    editor = "Riloff, Ellen  and
      Chiang, David  and
      Hockenmaier, Julia  and
      Tsujii, Jun{'}ichi",
    booktitle = "Proceedings of the 2018 Conference on Empirical Methods in Natural Language Processing",
    month = oct # "-" # nov,
    year = "2018",
    address = "Brussels, Belgium",
    publisher = "Association for Computational Linguistics",
    url = "https://aclanthology.org/D18-1325/",
    doi = "10.18653/v1/D18-1325",
    pages = "2947--2954"
}

@inproceedings{maruf-etal-2019-selective,
    title = "Selective Attention for Context-aware Neural Machine Translation",
    author = "Maruf, Sameen  and
      Martins, Andr{\'e} F. T.  and
      Haffari, Gholamreza",
    editor = "Burstein, Jill  and
      Doran, Christy  and
      Solorio, Thamar",
    booktitle = "Proceedings of the 2019 Conference of the North {A}merican Chapter of the Association for Computational Linguistics: Human Language Technologies, Volume 1 (Long and Short Papers)",
    month = jun,
    year = "2019",
    address = "Minneapolis, Minnesota",
    publisher = "Association for Computational Linguistics",
    url = "https://aclanthology.org/N19-1313/",
    doi = "10.18653/v1/N19-1313",
    pages = "3092--3102"
}

@article{liu-etal-2024-lost,
    title = "Lost in the Middle: How Language Models Use Long Contexts",
    author = "Liu, Nelson F.  and
      Lin, Kevin  and
      Hewitt, John  and
      Paranjape, Ashwin  and
      Bevilacqua, Michele  and
      Petroni, Fabio  and
      Liang, Percy",
    journal = "Transactions of the Association for Computational Linguistics",
    volume = "12",
    year = "2024",
    address = "Cambridge, MA",
    publisher = "MIT Press",
    url = "https://aclanthology.org/2024.tacl-1.9/",
    doi = "10.1162/tacl_a_00638",
    pages = "157--173"
}

@article{10.1162/tacl_a_00343,
    author = {Liu, Yinhan and Gu, Jiatao and Goyal, Naman and Li, Xian and Edunov, Sergey and Ghazvininejad, Marjan and Lewis, Mike and Zettlemoyer, Luke},
    title = {Multilingual Denoising Pre-training for Neural Machine Translation},
    journal = {Transactions of the Association for Computational Linguistics},
    volume = {8},
    pages = {726-742},
    year = {2020},
    month = {11},
    issn = {2307-387X},
    doi = {10.1162/tacl_a_00343},
    url = {https://doi.org/10.1162/tacl_a_00343},
    eprint = {https://direct.mit.edu/tacl/article-pdf/doi/10.1162/tacl_a_00343/1923401/tacl_a_00343.pdf},
}

@article{10.1162/tacl_a_00683,
    author = {Guerreiro, Nuno M. and Rei, Ricardo and Stigt, Daan van and Coheur, Luisa and Colombo, Pierre and Martins, André F. T.},
    title = {xcomet: Transparent Machine Translation Evaluation through Fine-grained Error Detection},
    journal = {Transactions of the Association for Computational Linguistics},
    volume = {12},
    pages = {979-995},
    year = {2024},
    month = {09},
    issn = {2307-387X},
    doi = {10.1162/tacl_a_00683},
    url = {https://doi.org/10.1162/tacl_a_00683},
    eprint = {https://direct.mit.edu/tacl/article-pdf/doi/10.1162/tacl_a_00683/2468704/tacl_a_00683.pdf},
}

@book{halliday1976cohesion,
  author    = {Halliday, M. A. K. and Hasan, Ruqaiya},
  title     = {Cohesion in English},
  edition   = {1st},
  publisher = {Routledge},
  address   = {London},
  year      = {1976},
  doi       = {10.4324/9781315836010}
}

@article{Hardmeier2012DiscourseIS,
  author  = {Hardmeier, Christian},
  title   = {Discourse in Statistical Machine Translation: A Survey and a Case Study},
  journal = {Discours},
  volume  = {11},
  year    = {2012},
  doi     = {10.4000/discours.8726},
  url     = {https://journals.openedition.org/discours/8726}
}

@misc{anthropic2025opus46,
  author       = {{Anthropic}},
  title        = {Claude {Opus} 4.6 System Card},
  year         = {2025},
  publisher    = {Anthropic},
  url          = {https://www.anthropic.com/claude-opus-4-6-system-card},
  note         = {Accessed: 2026-05-17}
}

@misc{google2026gemini3flash,
  author       = {{Google DeepMind}},
  title        = {Gemini 3 Flash},
  year         = {2026},
  url = {https://ai.google.dev/gemini-api/docs/models/gemini-3-flash-preview},
  note         = {Accessed: 2026-05-16}
}

@misc{deepseekai2026deepseekv4,
      title={DeepSeek-V4: Towards Highly Efficient Million-Token Context Intelligence},
      author={{DeepSeek-AI}},
      year={2026},
      url={https://api-docs.deepseek.com/quick_start/pricing},
}

@misc{post2024escapingsentencelevelparadigmmachine,
      title={Escaping the sentence-level paradigm in machine translation}, 
      author={Matt Post and Marcin Junczys-Dowmunt},
      year={2024},
      eprint={2304.12959},
      archivePrefix={arXiv},
      primaryClass={cs.CL},
      url={https://arxiv.org/abs/2304.12959}, 
}

@article{cleveland1981lowess,
  author  = {Cleveland, William S.},
  title   = {{LOWESS}: A Program for Smoothing Scatterplots by Robust Locally Weighted Regression},
  journal = {The American Statistician},
  volume  = {35},
  number  = {1},
  pages   = {54},
  year    = {1981},
  doi     = {10.2307/2683591}
}

@incollection{clark1977comprehension,
  author    = {Clark, Herbert H. and Haviland, Susan E.},
  title     = {Comprehension and the Given-New Contract},
  booktitle = {Discourse Production and Comprehension},
  editor    = {Freedle, Roy O.},
  series    = {Discourse Processes: Advances in Research and Theory},
  volume    = {1},
  chapter   = {1},
  pages     = {1--40},
  publisher = {Ablex Publishing Corporation},
  address   = {Norwood, NJ},
  year      = {1977},
  url       = {http://www.web.stanford.edu/~clark/1970s/Clark,%20H.H.%20_%20Haviland,%20S.E.%20_Comprehension%20and%20the%20given-new%20contract_%201977.pdf}
}

@incollection{Prince1981TowardAT,
  address = {New York},
  author = {Prince, Ellen F.},
  booktitle = {Syntax and semantics: Vol. 14. Radical Pragmatics},
  editor = {Cole, P.},
  pages = {223--255},
  publisher = {Academic Press},
  title = {Toward a taxonomy of given-new information},
  year = 1981
}

@article{HOBBS1978311,
title = {Resolving pronoun references},
journal = {Lingua},
volume = {44},
number = {4},
pages = {311-338},
year = {1978},
issn = {0024-3841},
doi = {https://doi.org/10.1016/0024-3841(78)90006-2},
url = {https://www.sciencedirect.com/science/article/pii/0024384178900062},
author = {Jerry R. Hobbs}
}

@article{10.5555/12457.12458,
    author = {Grosz, Barbara J. and Sidner, Candace L.},
    title = {Attention, intentions, and the structure of discourse},
    year = {1986},
    issue_date = {July-September 1986},
    publisher = {MIT Press},
    address = {Cambridge, MA, USA},
    volume = {12},
    number = {3},
    issn = {0891-2017},
    journal = {Comput. Linguist.},
    month = jul,
    pages = {175–204},
    numpages = {30},
    url={https://dl.acm.org/doi/10.5555/12457.12458},
}

@inproceedings{zouhar-etal-2026-generating,
    title = "Generating Difficult-to-Translate Texts",
    author = "Zouhar, Vil{\'e}m  and
      Xu, Wenda  and
      Riley, Parker  and
      Juraska, Juraj  and
      Finkelstein, Mara  and
      Freitag, Markus  and
      Deutsch, Daniel",
    editor = "Chen, Pinzhen  and
      Zouhar, Vil{\'e}m  and
      Hu, Hanxu  and
      Khanuja, Simran  and
      Zhu, Wenhao  and
      Haddow, Barry  and
      Birch, Alexandra  and
      Aji, Alham Fikri  and
      Sennrich, Rico  and
      Hooker, Sara",
    booktitle = "Proceedings of the First Workshop on Multilingual Multicultural Evaluation",
    month = mar,
    year = "2026",
    address = "Rabat, Morocco",
    publisher = "Association for Computational Linguistics",
    url = "https://aclanthology.org/2026.mme-main.14/",
    doi = "10.18653/v1/2026.mme-main.14",
    pages = "204--219",
    ISBN = "979-8-89176-368-5"
}

@article{10.1162/TACL.a.60,
    author = {Zouhar, Vilém and Cui, Peng and Sachan, Mrinmaya},
    title = {How to Select Datapoints for Efficient Human Evaluation of NLG Models?},
    journal = {Transactions of the Association for Computational Linguistics},
    volume = {13},
    pages = {1789-1811},
    year = {2025},
    month = {12},
    issn = {2307-387X},
    doi = {10.1162/TACL.a.60},
    url = {https://doi.org/10.1162/TACL.a.60},
    eprint = {https://direct.mit.edu/tacl/article-pdf/doi/10.1162/TACL.a.60/2571151/tacl.a.60.pdf},
}

@misc{akhtar2026aibenchmarksplateausystematic,
      title={When AI Benchmarks Plateau: A Systematic Study of Benchmark Saturation}, 
      author={Mubashara Akhtar and Anka Reuel and Prajna Soni and Sanchit Ahuja and Pawan Sasanka Ammanamanchi and Ruchit Rawal and Vilém Zouhar and Srishti Yadav and Chenxi Whitehouse and Dayeon Ki and Jennifer Mickel and Leshem Choshen and Marek Šuppa and Jan Batzner and Jenny Chim and Jeba Sania and Yanan Long and Hossein A. Rahmani and Christina Knight and Yiyang Nan and Jyoutir Raj and Yu Fan and Shubham Singh and Subramanyam Sahoo and Eliya Habba and Usman Gohar and Siddhesh Pawar and Robert Scholz and Arjun Subramonian and Jingwei Ni and Mykel Kochenderfer and Sanmi Koyejo and Mrinmaya Sachan and Stella Biderman and Zeerak Talat and Avijit Ghosh and Irene Solaiman},
      year={2026},
      eprint={2602.16763},
      archivePrefix={arXiv},
      primaryClass={cs.AI},
      url={https://arxiv.org/abs/2602.16763}, 
}

@misc{gemma4,
  title        = {Gemma 4: Byte for byte, the most capable open models},
  author       = {{Gemma Team}},
  year         = {2026},
  month        = {April},
  howpublished = {Google DeepMind Blog},
  url          = {https://blog.google/innovation-and-ai/technology/developers-tools/gemma-4/}
}

@misc{singh2026openaigpt5card,
      title={OpenAI GPT-5 System Card}, 
      author={Aaditya Singh and Adam Fry and Adam Perelman and others},
      year={2026},
      eprint={2601.03267},
      archivePrefix={arXiv},
      primaryClass={cs.CL},
      url={https://arxiv.org/abs/2601.03267}, 
}

@misc{navercloudhyperclovaxteam2026hyperclovax32bthink,
      title={HyperCLOVA X 32B Think}, 
      author={NAVER, Cloud HyperCLOVA X Team},
      year={2026},
      eprint={2601.03286},
      archivePrefix={arXiv},
      primaryClass={cs.CV},
      url={https://arxiv.org/abs/2601.03286}, 
}

@misc{qwen3.5,
    title  = {{Qwen3.5}: Towards Native Multimodal Agents},
    author = {{Qwen Team}},
    month  = {February},
    year   = {2026},
    url    = {https://qwen.ai/blog?id=qwen3.5}
}

@inproceedings{raunak:23-leveraging,
    title = "Leveraging {GPT}-4 for Automatic Translation Post-Editing",
    author = "Raunak, Vikas  and
      Sharaf, Amr  and
      Wang, Yiren  and
      Awadalla, Hany  and
      Menezes, Arul",
    editor = "Bouamor, Houda  and
      Pino, Juan  and
      Bali, Kalika",
    booktitle = "Findings of the Association for Computational Linguistics: EMNLP 2023",
    month = dec,
    year = "2023",
    address = "Singapore",
    publisher = "Association for Computational Linguistics",
    url = "https://aclanthology.org/2023.findings-emnlp.804/",
    doi = "10.18653/v1/2023.findings-emnlp.804",
    pages = "12009--12024"
}

\newpage
\appendix
\section*{Appendix}

\crefalias{section}{appendix}
\crefalias{subsection}{appendix}
\crefalias{subsubsection}{appendix}

\begin{table*}[t]
\centering
\small
\setlength{\tabcolsep}{4pt}
\begin{tabular}{lcccc|cccc|cccc}
\toprule
\multirow{2}{*}{\textbf{Domain}} & \multicolumn{4}{c}{\textbf{spaCy} (\texttt{trf})} & \multicolumn{4}{c}{\textbf{Flair}} & \multicolumn{4}{c}{\textbf{spaCy} (\texttt{sm}, weak)} \\
\cmidrule(lr){2-5}\cmidrule(lr){6-9}\cmidrule(lr){10-13}
 & $\mu$ & Med & Max & \%$=$0 & $\mu$ & Med & Max & \%$=$0 & $\mu$ & Med & Max & \%$=$0 \\
\midrule
News     &  9.7 &  4 &  52 & 28.9 &  9.2 &  4 &  50 & 32.9 & 11.2 &  4 &  50 & 26.2 \\
Social   &  3.8 &  0 & 166 & 62.4 &  4.1 &  0 & 130 & 63.0 &  4.4 &  0 &  82 & 60.9 \\
Literary & 52.8 & 44 & 171 & 13.1 & 52.2 & 33 & 200 & 16.0 & 44.6 & 32 & 162 & 15.5 \\
\bottomrule
\end{tabular}
\caption{Per-domain \ddp\ statistics under spaCy (\texttt{en\_core\_web\_trf}), Flair, and a deliberately weaker extractor (\texttt{en\_core\_web\_sm}). The Literary $>$ News $>$ Social ordering and the ordering of $\ddpop{=}0$ shares are preserved under all three, including the weakened pipeline.}
\label{tab:ablation_stats}
\end{table*}

\section{\ddp Computation}

\subsection{Validation against Gold Coreference}
\label{app:ontonotes}

To assess how much of true inter-sentential dependency our two-anchor formulation captures, we compare DDP against gold coreference chains from OntoNotes 5.0 \citep{hovy-etal-2006-ontonotes}. We select 80 English documents from the news ($n{=}45$) and narrative ($n{=}35$) portions whose genre overlaps with our evaluation data, yielding 2{,}143 segments.

\paragraph{Gold-distance computation.}
For each segment $s_i$, we extract all mention spans whose coreference chain has at least one antecedent in a prior segment $s_j$ ($j < i$). The gold dependency distance is defined symmetrically to DDP:
\begin{equation}
d_{\text{gold}}(s_i) \;=\; \max_{m \in M_i}\,(i - j_m),
\end{equation}
where $M_i$ is the set of mentions in $s_i$ with a prior antecedent and $j_m$ is the segment index of the most distant reachable chain link. Segments without prior-chain mentions receive $d_{\text{gold}} = 0$.

\paragraph{Agreement.}
Segment-level DDP correlates with gold distance at Pearson $r = 0.81$ (Spearman $\rho = 0.78$), with stronger agreement on news ($r = 0.86$) than narrative ($r = 0.74$). The narrative gap reflects the denser presence of bridging anaphora and lexical chains, which fall outside our anchor inventory.

\paragraph{One-sided error.}
Consistent with its construction, DDP underestimates gold distance in $86.4\%$ of segments and matches it in $12.8\%$. Overestimation occurs in $0.8\%$ and is fully attributable to NER false positives that introduce spurious early mentions. This confirms the lower-bound guarantee in practice.

\paragraph{Implications.}
Both empirical claims in this paper, that current benchmarks undersample high-dependency segments and that human-metric divergence grows with them, depend on the presence rather than the exhaustive enumeration of high-dependency cases. Since DDP only under-counts, segments it identifies as high-dependency are also high-distance under gold annotation. The moderate gap on narrative further suggests that the divergence pattern may be \emph{underestimated} by DDP-stratified analysis, as some segments treated as low-DDP carry hidden bridging or lexical-chain dependencies that would reinforce the observed effect if measured. Extending the anchor inventory to bridging and lexical cohesion is a natural direction for future work.

\subsection{Worked examples on WMT24++}
We illustrate \ddp\ computation on three documents from WMT24++, one per domain in \Cref{fig:ddp_example}. Documents are chosen to display the operation of both anchor rules, and their \ddp\ statistics are close to or above the respective domain means, allowing the figures to show meaningful anchor activity without depicting extreme tails.

\begin{figure}
    \centering
    \includegraphics[width=0.8\linewidth]{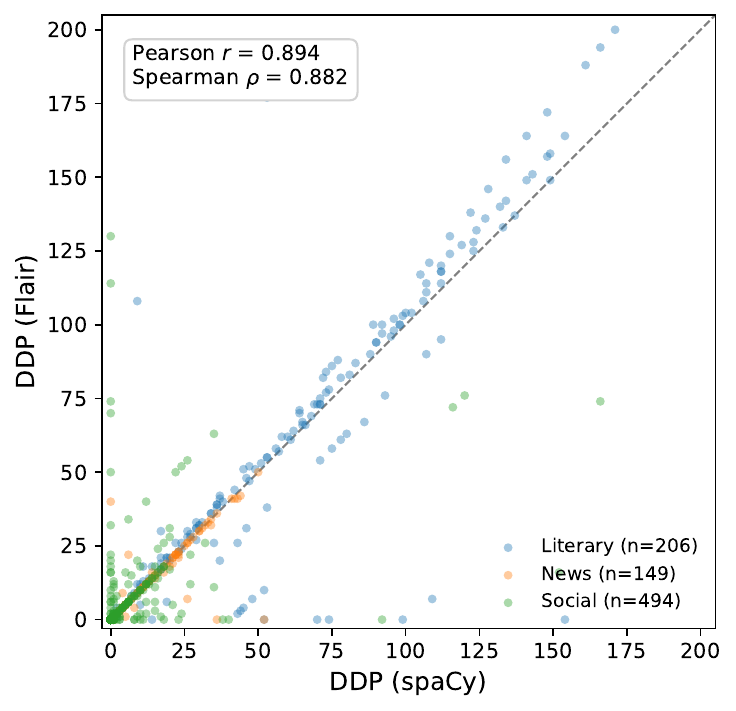}
    \caption{Segment-lveel \ddp from spaCy vs Flair on WMT24++. Each point is a unique source segment, colored by domain. Most segmetns lie on or near the identity line.}
    \label{fig:ablation_scatter}
\end{figure}

\subsection{Ablation: Anchor Extraction Tool}
\label{appx:ablation}

We replicate \ddp\ with Flair NER and POS tagging \citep{akbik-etal-2019-flair} in place of spaCy. Sentence segmentation, dependency parsing for head-noun extraction, and downstream distance computation are held constant to isolate the effect of the NER and POS components.

\paragraph{Agreement on \ddp\ values.}
\Cref{fig:ablation_scatter} plots segment-level \ddp\ from the two pipelines against each other. The values are strongly correlated (Pearson $r=0.89$, Spearman $\rho=0.88$), with 69\% of segments receiving identical \ddp\ and 85\% agreeing within $\pm5$. Disagreement concentrates in segments with longer referential chains, where small differences in entity recognition propagate through chain accumulation.

\begin{table}[t]
\small
\centering
\begin{tabular}{llrrrr}
\toprule
& & \multicolumn{2}{c}{\textbf{WMT24++}} 
& \multicolumn{2}{c}{\textbf{WMT25}} \\
\cmidrule(lr){3-4}\cmidrule(lr){5-6}
\textbf{Domain} & & $\mu$ & max 
                & $\mu$ & max \\
\midrule
News     &&   9.7 &  52 &  10.6 &  41 \\
Social   &&   3.8 & 166 &  21.3 & 186 \\
Literary &&  52.8 & 171 & 222.6 & 482 \\
\bottomrule
\end{tabular}
\caption{Mean and maximum segment-level \ddp\ scores across domains in WMT24++ and WMT25.}
\label{tab:ddp_stats}
\end{table}

\begin{figure*}[t]
    \centering
    \includegraphics[width=1\linewidth]{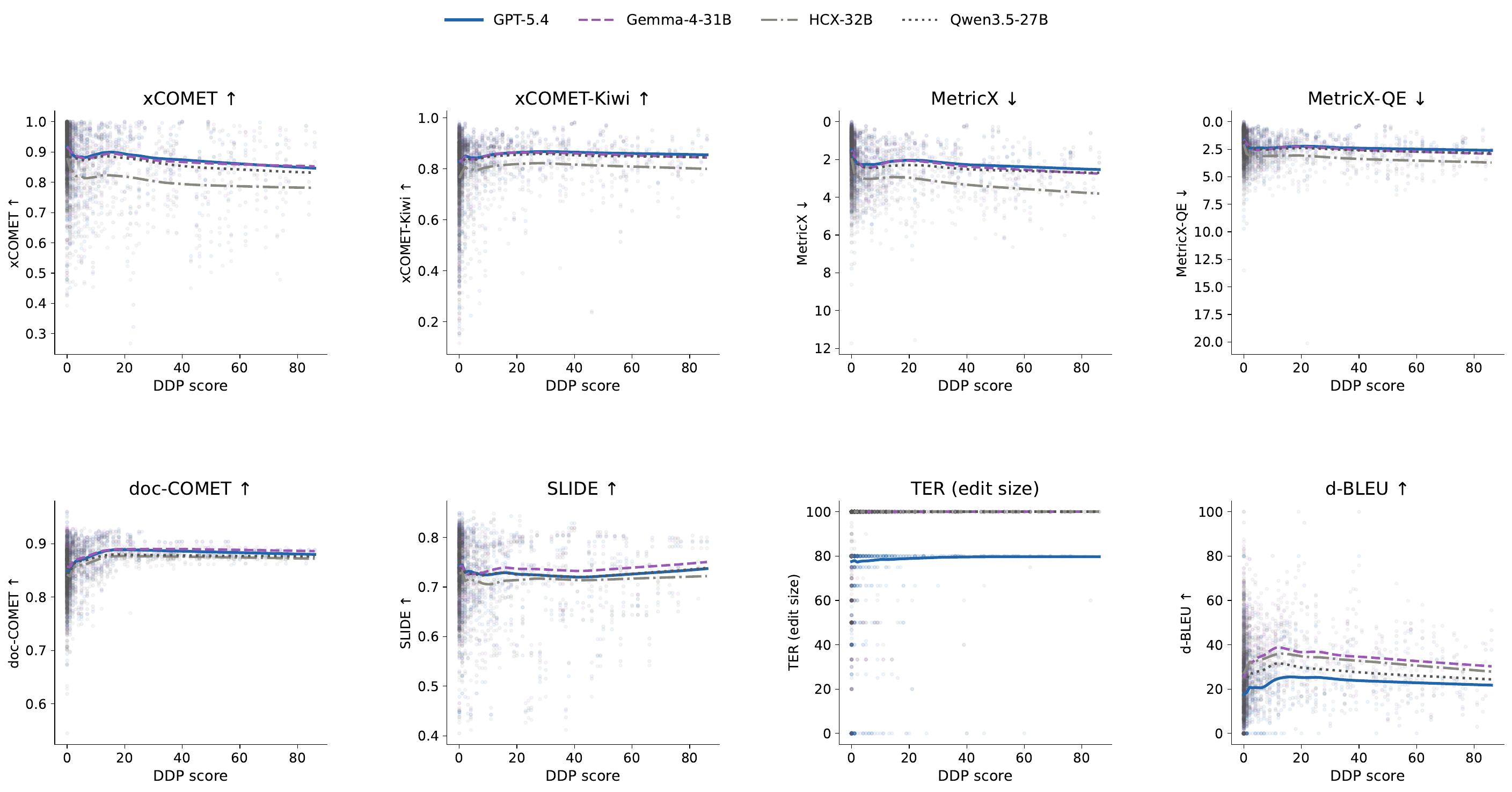}
    \caption{LOWESS-smoothed metric scores per model as a function of segment-level \ddp\ (\textbf{En--Zh}, all injection strategies averaged; four models available). As in En--Ko (\Cref{fig:ddp_model}), metric scores remain largely flat across the \ddp\ spectrum and no model shows consistent improvement or degradation as discourse dependency increases, confirming that the insensitivity of automatic metrics to \ddp\ is not specific to Korean.}
    \label{fig:enzh}
\end{figure*}

\paragraph{Domain-level statistics.}
\Cref{tab:ablation_stats} reports per-domain \ddp\ under both tools. Domain means shift by less than one point in absolute terms, and the ordering Literary $>$ News $>$ Social is preserved. The qualitative findings of \Cref{subsec:wmt_analysis}, that benchmarks skew toward low-\ddp\ segments and that domain labels conflate within-domain variation, hold under both tools.

\paragraph{A deliberately weaker tagger.}
Agreement between two strong taggers leaves open whether the ordering depends on tagger quality. We therefore repeat the replication with \texttt{en\_core\_web\_sm}, holding sentence segmentation and distance computation constant. Tagger quality affects the two rules asymmetrically. Under Rule~\circled{1}, a missed mention can only shorten a chain, since a first mention cannot be invented. Under Rule~\circled{2}, a pronoun is anchored to the \emph{most recent} compatible antecedent, so a dropped intervening entity moves the antecedent further back and lengthens the distance. Neither error produces the Literary $>$ News $>$ Social ordering, though individual segments may move in either direction.

The ordering is preserved (\Cref{tab:ablation_stats}). Segment-level agreement with spaCy is high (Pearson $r=0.84$, Spearman $\rho=0.88$), with 73.4\% exact agreement and 83.8\% within $\pm 5$, and the ordering of $\ddpop{=}0$ shares is also retained. Literary means fall (52.8 $\rightarrow$ 44.6), where long chains offer the most links to miss, while news and social rise slightly (9.7 $\rightarrow$ 11.2, 3.8 $\rightarrow$ 4.4) through the Rule~\circled{2} effect. \ddp\ decreases on 14.7\% of segments and increases on 11.9\%. The lower bound of \Cref{subsec:operation} therefore holds strictly for entity re-mentions and approximately for pronominal chains, which does not affect the domain-level results but bears on certifying individual segments as high-dependency.

\subsection{Cross-lingual Consistency: En--Zh}
\label{appx:enzh}
Since \ddp\ is computed from the English source, the same scores apply to any target language. We report automatic metric results for En--Zh on the four models for which outputs were available (\mGPT, \mGemma, \mQwen, \mHCX). Metric scores remain largely flat across the \ddp\ spectrum as in En--Ko (\Cref{fig:enzh}), so metric insensitivity to discourse dependency is not specific to Korean. \mHCX\ ranks lower here than in En--Ko, consistent with its Korean-specialized training.

Across injection strategies (\Cref{fig:ddp_metrics_enzh}), most metrics again fail to separate them, and the gap between APE and \HPE\ is smaller than in En--Ko. Two readings are available. The evaluated models, \mHCX\ aside, have stronger Chinese coverage, which would narrow the true APE--\HPE\ gap below what current metrics can resolve; alternatively, Chinese discourse properties such as pervasive zero-anaphora and topic-drop may leave fewer surface traces for metrics to detect. Distinguishing the two requires human evaluation, and whether human preferences follow the same \ddp-stratified pattern across language pairs remains open.
\newpage
\section{Experimental Details}

\subsection{Dataset statistics}
\label{appx:data_stat}
We use the English-Korean subset of WMT24++ \citep{deutsch-etal-2025-wmt24} for all experiments. The released data spans three domains (literary, news, social) across 59 documents and 849 unique source segments. \Cref{tab:data_stats} reports the per-domain breakdown along with \ddp\ statistics computed on the English source. The three domains differ sharply in their dependency profiles: 62.3\% of social segments have $\ddpop = 0$ compared with 13.1\% of literary segments, and the literary mean (52.8) is over an order of magnitude larger than the social mean (3.8). News falls between the two extremes (mean 9.7, 28.9\% at $\ddpop=0$). All documents in the subset contain at least two segments, so no further filtering is applied.

\subsection{Context Injection Strategy: Construction}
\label{appx:ctx_construct}

\paragraph{\APErel.}
We encode all source segments $\{s_j\}_{j \neq i}$ in each document using LaBSE \citep{feng-etal-2022-language}, a multilingual sentence encoder that produces language-agnostic representations. The current segment $s_i$ is excluded from the index to prevent self-retrieval. Given $s_i$ as query, we select the top-$k$ segments by cosine similarity and pair them with their target segments. Retrieved pairs are reordered by their original document position before being injected into the prompt, preserving sequential structure across conditions.

\paragraph{\APEexp.}
For each document, we prompt \mClaude \citep{anthropic2025opus46} to extract structured annotations covering three discourse dimensions that affect target-language editing decisions:
\begin{itemize}[noitemsep]
    \item \textbf{Genre and domain}: text type and subject domain (e.g., \textit{news report}, \textit{literary dialogue}).
    \item \textbf{Participant relationships}: roles and social relationships between discourse participants where identifiable (e.g., \textit{interviewer-interviewee}, \textit{narrator-character}).
    \item \textbf{Register and formality}: expected formality level and register conventions, such as honorific speech level choices in Korean and written vs.\ colloquial distinctions in Chinese.
\end{itemize}

\begin{table}[t]
\centering
\small
\setlength{\tabcolsep}{4pt}
\resizebox{\columnwidth}{!}{
\begin{tabular}{lrrrrrrr}
\toprule
\multirow{2}{*}{\textbf{Domain}} & \multirow{2}{*}{\textbf{Docs}} & \multirow{2}{*}{\textbf{Segs}} & \multicolumn{4}{c}{$\ddpop$} & \multirow{2}{*}{\textbf{\%$=\!0$}} \\
\cmidrule(lr){4-7}
 & & & mean (std) & med & p95 & max & \\
\midrule
News     & 17 & 149 &  9.7 (12.4) &  4 &  35 &  52 & 28.9 \\
Social   & 34 & 494 &  3.8 (14.3) &  0 &  17 & 166 & 62.3 \\
Literary &  8 & 206 & 52.8 (46.9) & 44 & 141 & 171 & 13.1 \\
\midrule
Total    & 59 & 849 & 16.7 (33.2) &  1 & 101 & 171 & 44.5 \\
\bottomrule
\end{tabular}
}
\caption{Statistics of the WMT24++ En--Ko subset used in our experiments. \textbf{\%$=\!0$} reports the share of segments with $\ddpop=0$, for which no prior discourse context is required. Social texts skew strongly toward self-contained segments, while literary texts span a wide range of dependencies.}
\label{tab:data_stats}
\end{table} 

Annotations are produced in two stages. We first prompt \mClaude on the full source document to generate a JSON annotation per document via greedy decoding. It is not among the evaluated models, avoiding circular dependency between annotation and evaluation. A trained annotator then reviews each annotation and corrects errors, ambiguities, and cases where the model fails to identify participant relationships or register conventions. The annotation guidelines and inference prompt are in \Cref{fig:exp_prompt}.

\subsection{Context Injection Prompts}
\label{appx:prompt}

A unified prompt structure is used across all conditions (\Cref{fig:prompt}). The system prompt is identical and contains only a role definition, so any performance difference reflects context content rather than prompt formulation. All task instructions and quality guidelines are placed in the user prompt, where they are processed jointly with the provided context.

The context block is the only element that varies across conditions. \APEseg receives no context. \APEseq, \APErel, and \APEfull each receive a source and target context block populated according to the curation strategy. \APEexp receives a structured document-information block in place of raw context, with the register instruction adjusted to signal that the relevant information is supplied explicitly rather than requiring inference. No instruction is given regarding the relevance or quality of the provided context, so the model's sensitivity to context content can be observed without interference.

\subsection{Pilot Study: Choice of $k$}
\label{appx:pilot}

The number of retrieved segments $k$ controls the amount of context injected in \APEseq and \APErel. To determine an appropriate value, we conduct a pilot study on a held-out subset of ${\approx}$270 segments from WMT25 En--Ko \citep{kocmi-etal-2025-findings}, using \mGemma \citep{gemma4} as the primary model with $k \in \{1, 3, 5, 10\}$.

We select $k$ based on two criteria. First, we measure the divergence between \APEseq\ and \APErel\ via \ter:
\begin{equation}
    \Delta\text{\ter}(k) =
    \left| \text{\ter}_{\text{seq}} - 
    \text{\ter}_{\text{rel}} \right|
\end{equation}
A larger $\Delta\text{\ter}(k)$ indicates that the two injection strategies produce meaningfully different outputs, exposing the contrast between sequential and relevance-based selection. Second, we track the generation failure rate, 
defined as the proportion of segments with \ter~$>$~100.

We select the $k$ that maximizes $\Delta\text{\ter}(k)$ subject to a low failure rate. As shown in \Cref{fig:determining_k}, $k{=}10$ yields the largest divergence but also the highest failure rate, making it unreliable. $k{=}5$ achieves the best trade-off: sufficient divergence between strategies with the lowest hallucination rate across all settings. We therefore fix $k{=}5$ for all subsequent experiments.

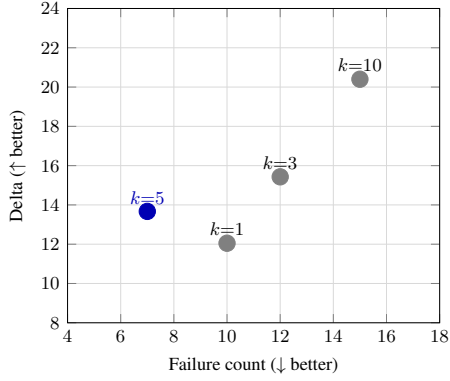
\begin{figure}[t]
\centering
\resizebox{0.8\linewidth}{!}{
\begin{tikzpicture}
\begin{axis}[
    width=8cm,
    height=7cm,
    xlabel={Failure count ($\downarrow$ better)},
    ylabel={Delta ($\uparrow$ better)},
    xlabel style={font=\small},
    ylabel style={font=\small},
    tick label style={font=\small},
    xmin=4,  xmax=18,
    ymin=8,  ymax=24,
    xtick={4,6,8,10,12,14,16,18},
    ytick={8,10,12,14,16,18,20,22,24},
    grid=both,
    grid style={line width=0.3pt, draw=gray!30},
    clip=false,
]

\addplot[only marks, mark=*, mark size=4pt,
    mark options={fill=gray, draw=gray}]
    coordinates {(10, 12.05)};
\node[above, font=\small] at (axis cs:10, 12.05) {$k{=}1$};

\addplot[only marks, mark=*, mark size=4pt,
    mark options={fill=gray, draw=gray}]
    coordinates {(12, 15.43)};
\node[above, font=\small] at (axis cs:12, 15.43) {$k{=}3$};

\addplot[only marks, mark=*, mark size=4pt,
    mark options={fill=blue!70!black, draw=blue!70!black}]
    coordinates {(7, 13.67)};
\node[above, font=\small\bfseries, text=blue!70!black]
    at (axis cs:7, 13.67) {$k{=}5$};

\addplot[only marks, mark=*, mark size=4pt,
    mark options={fill=gray, draw=gray}]
    coordinates {(15, 20.40)};
\node[above, font=\small] at (axis cs:15, 20.40) {$k{=}10$};

\end{axis}
\end{tikzpicture}
}
\caption{Delta vs.\ Failure count per $k$. $k{=}5$ (\textcolor{blue}{blue}) achieves the lowest failure count with moderate delta.}
\label{fig:determining_k}
\end{figure} 

\subsection{Human Evaluation Details}
\label{appx:humeval}

\paragraph{Sampling Criteria.}
To ensure that human evaluation focuses on documents where context curation produces measurable variation, we select 
documents based on inter-condition divergence rather than uniform random sampling. For each document, we compute the mean pairwise TER distance among all curation strategy outputs, excluding segments with \ter~$>$~100 as generation failures \citep{raunak:23-leveraging}. Documents are ranked by mean divergence in descending order, and the top-$k$ documents per domain are retained as candidates. The final subset is stratified by domain and capped at ${\approx}$500 segments per assumption, yielding 1{,}000 segments in total, with no overlap between \Aone\ and \Atwo\ subsets. Within each selected document, APE outputs are anonymized and presented in randomized order to mitigate position bias.

\paragraph{Annotation Platform.}
We use Label Studio as the annotation platform.\footnote{\url{https://labelstud.io}} Each task presents the current segment alongside its immediately preceding and following source segments for discourse context. Annotators may freely navigate between segments within the same document for reference and may revise their rankings at any time before submission.

\paragraph{Annotators.}
We recruit four professional translators with at least three years of experience in En-Ko translation. Each annotator evaluated ${\approx}$500 segments. Annotators were compensated at a competitive industry rate and consented 
to the release of their annotations.

\paragraph{Annotation Guidelines.}
Annotators were instructed to rank the four candidate translations from 1 (best) to 4 (worst) based on overall translation quality, with ties permitted when candidates were judged indistinguishable. They were asked to consider not only fluency and adequacy, but also discourse attributes, and were encouraged to consult surrounding segments when assessing referential coherence.

\paragraph{Inter-Annotator Agreement.}
IAA is measured using Kendall's $\tau$ computed between the two independent rankings per segment, averaged across all evaluated segments. The overall $\tau$ was 0.37, consistent with prior relative ranking evaluations in MT \citep{callison-burch-etal-2007-meta}, and indicates moderate agreement.

\paragraph{Remuneration.}
Professional translators are compensated at competitive industry rates commensurate with their expertise and time. All annotators provided informed consent for the use and publication of their annotations.

\begin{table*}[!ht]
\small
\centering
\setlength{\tabcolsep}{5pt}
\resizebox{0.8\linewidth}{!}{
\begin{tabular}{llrrrrrr}
\toprule
\textbf{Model} & \textbf{Strategy} &
\textbf{Input} & \textbf{Input} &
\textbf{Output} &
\textbf{Latency} & \textbf{Latency} \\
& & \textbf{(tok)} & \textbf{(ratio)} &
\textbf{(tok)} &
\textbf{(sec)} & \textbf{(ratio)} \\
\midrule
    \mGPT & \APEseg & 216 & 1.0$\times$ & 68 & -- & -- \\
     & \APEseq & 644 & 3.0$\times$ & 60 & -- & -- \\
     & \APErel & 730 & 3.4$\times$ & 62 & -- & -- \\
     & \APEexp & 347 & 1.6$\times$ & 72 & -- & -- \\
     & \APEfull & 10,688 & 49.5$\times$ & 62 & -- & -- \\
    \midrule
    \mGemini & \APEseg & 208 & 1.0$\times$ & 32 & -- & -- \\
     & \APEseq & 636 & 3.1$\times$ & 32 & -- & -- \\
     & \APErel & 718 & 3.5$\times$ & 32 & -- & -- \\
     & \APEexp & 343 & 1.6$\times$ & 32 & -- & -- \\
     & \APEfull & 17,556 & 84.4$\times$ & 34 & -- & -- \\
    \midrule
    \mDeepSeek & \APEseg & 218 & 1.0$\times$ & 880 & -- & --  \\
     & \APEseq & 686 & 3.1$\times$ & 927 & -- & --  \\
     & \APErel & 781 & 3.6$\times$ & 952 & -- & --  \\
     & \APEexp & 364 & 1.7$\times$ & 934 & -- & --  \\
     & \APEfull & 12,080 & 55.4$\times$ & 906 & -- & --  \\
    \midrule
    \mGemma & \APEseg & 243 & 1.0$\times$ & 57 & 2.90 & 1.00$\times$ \\
     & \APEseq & 665 & 2.7$\times$ & 54 & 3.26 & 1.12$\times$ \\
     & \APErel & 747 & 3.1$\times$ & 56 & 3.40 & 1.17$\times$ \\
     & \APEexp & 372 & 1.5$\times$ & 57 & 3.27 & 1.13$\times$ \\
     & \APEfull & 10,371 & 42.7$\times$ & 57 & 3.89 & 1.34$\times$ \\
    \midrule
    \mHCX & \APEseg & 220 & 1.0$\times$ & 47 & 4.09 & 1.00$\times$ \\
     & \APEseq & 613 & 2.8$\times$ & 44 & 7.45 & 1.82$\times$ \\
     & \APErel & 690 & 3.1$\times$ & 46 & 7.90 & 1.93$\times$ \\
     & \APEexp & 343 & 1.6$\times$ & 46 & 2.81 & 0.69$\times$ \\
     & \APEfull & 8,947 & 40.7$\times$ & 49 & 13.18 & 3.22$\times$ \\
    \midrule
    \mQwen & \APEseg & 222 & 1.0$\times$ & 53 & 2.49 & 1.00$\times$ \\
     & \APEseq & 631 & 2.8$\times$ & 51 & 2.79 & 1.12$\times$ \\
     & \APErel & 708 & 3.2$\times$ & 53 & 2.95 & 1.18$\times$ \\
     & \APEexp & 358 & 1.6$\times$ & 54 & 2.60 & 1.04$\times$ \\
     & \APEfull & 9,471 & 42.7$\times$ & 56 & 12.33 & 4.95$\times$ \\
\bottomrule
\end{tabular}
}
\caption{Inference overhead per model and curation strategy, averaged over all evaluation segments (En--Ko). Input/latency ratios are relative to \APEseg\ (null baseline). Latency for API-based models (\mGPT, \mGemini, \mDeepSeek) includes network round-trip time and is not directly comparable to open-weight models (reported as~--). \APEfull\ requires 40--84$\times$ more input tokens than \APEseg, yet provides no consistent improvement in translation quality (\Cref{sec:results}).}
\label{tab:efficiency}
\end{table*}

\subsection{Inference Overhead}
\label{appx:efficiency}

To assess the computational overhead of each context injection strategy, we record three efficiency metrics for every model--condition pair: (1) \textbf{input tokens}, which vary directly with context size and injection strategy; (2) \textbf{output tokens}, reflecting editing verbosity; and (3) \textbf{latency}, measured as wall-clock time per segment from prompt submission to response completion.

Input token counts are deterministic given the injection strategy and document, and are reported as averages over the full evaluation set. Latency is measured on a single NVIDIA RTX PRO 6000 Blackwell GPU under a fixed batch size of one to isolate per-segment processing time, averaged over three runs to reduce variance. API-based models (\mGPT, \mGemini, \mDeepSeek) include network round-trip time and are reported separately from open-weight models to avoid confounding infrastructure differences. Models that do not support a \texttt{seed} parameter are noted where applicable. Results for these models may exhibit minor run-to-run variation.

\Cref{tab:efficiency} shows that \APEfull\ incurs the highest input token cost across all models, requiring 40 - 84$\times$ more tokens than \APEseg, yet yields no consistent improvement in translation quality (\Cref{sec:results}). \APEexp\ is the most token-efficient context strategy (${\approx}$1.6$\times$ overhead), as structured declarative knowledge is considerably shorter than raw segments, but it consistently underperforms \APEseq\ in human evaluation. \APEseq\ and \APErel\ incur comparable input overhead (${\approx}$3$\times$), with \APErel\ marginally higher due to embedding-based retrieval of longer segments.

In terms of latency, open-weight models show modest increases from \APEseg\ to \APEseq\ (${\approx}$1.1 - 1.8$\times$), while \APEfull\ causes substantially higher latency on some models (up to 5$\times$ for \mQwen). Notably, \mDeepSeek generates substantially more output tokens than other models (880 - 952 vs.\ 32 - 72), likely reflecting tokenizer differences in Korean text generation. Taken together, \APEseq\ offers the best trade-off between context effectiveness and computational cost.

\begin{figure*}[t]
\centering
\scriptsize
\renewcommand{\arraystretch}{0.85}
\setlength{\tabcolsep}{4pt}

\begin{tabular}{p{0.04\linewidth} p{0.06\linewidth} p{0.84\linewidth}}
\toprule
\textbf{ID} & \textbf{$\ddpop$} & \textbf{(a) Literary} \textit{(Forever Snow)} \\
\midrule
871 & 0 & Prologue \\
\midrule
872 & 2 & 
"Please!" \textcolor{green!60!black}{Queen Eirwen}$^{[0]}$ begged, "don't send \textcolor{blue}{her}$^{[\to 1]}$ away! \textcolor{blue}{She}'s$^{[\to 1]}$ just a child!" \textcolor{green!60!black}{The Ice King}$^{[0]}$ looked down at \textcolor{blue}{his}$^{[\to 1]}$ wife from \textcolor{blue}{his}$^{[\to 1]}$ throne, \textcolor{blue}{his}$^{[\to 1]}$ royal advisor next to \textcolor{blue}{him}$^{[\to 1]}$. \textcolor{blue}{She}$^{[\to 2]}$ had given birth to a girl. \textcolor{blue}{He}$^{[\to 2]}$ scowled at the thought. \\
\midrule
873 & 6 & 
\textcolor{green!60!black}{The girl}$^{[0]}$ in question looked a lot like \textcolor{blue}{her}$^{[\to 1]}$ mother; pale blue eyes and white hair decorated with blue strands. \textcolor{green!60!black}{The Ice King}$^{[4]}$ thought a daughter made \textcolor{blue}{him}$^{[\to 5]}$ look weak. After all, \textcolor{blue}{he}$^{[\to 6]}$ had only had sons up until this point. \\
\midrule
874 & 11 &
"The only options you have are to kill the child or send \textcolor{blue}{it}$^{[\to 1]}$ out to the wilderness. This is \textcolor{blue}{her}$^{[\to 4]}$ punishment for your wrongdoing, \textcolor{green!60!black}{Eirwen}$^{[9]}$." \textcolor{blue}{He}$^{[\to 10]}$ spat out \textcolor{blue}{her}$^{[\to 5]}$ name as if \textcolor{blue}{it}$^{[\to 2]}$ was poison. \textcolor{blue}{He}$^{[\to 11]}$ always did. \\
\midrule
876 & 6 &
The \textcolor{green!60!black}{South}$^{[\to 2]}$ was desperate, and \textcolor{blue}{their}$^{[\to 2]}$ \textcolor{green!60!black}{royal family}$^{[0]}$ would do anything to stop the bloodshed. \textcolor{blue}{They}$^{[\to 2]}$ offered their only princess to be queen of the \textcolor{green!60!black}{North}$^{[\to 4]}$ and bear a son to be the future king. Then, the \textcolor{green!60!black}{South}$^{[\to 4]}$ would join with the \textcolor{green!60!black}{North}$^{[\to 5]}$ to make one kingdom. \textcolor{green!60!black}{The North}$^{[\to 6]}$ accepted, and \textcolor{green!60!black}{Princess Eirwen}$^{[0]}$ married \textcolor{green!60!black}{Prince Akull}$^{[0]}$. \\
\bottomrule
\end{tabular}

\vspace{0.5em}

\begin{tabular}{p{0.04\linewidth} p{0.06\linewidth} p{0.84\linewidth}}
\toprule
\textbf{ID} & \textbf{$\ddpop$} & \textbf{(b) News} \textit{(seattle\_times.800119, ``How to find out if you're flying on a Boeing 737 MAX'')} \\
\midrule
132 & 0 & How to find out if you're flying on a \textcolor{green!60!black}{Boeing 737 MAX}$^{[0]}$ \\
\midrule
133 & 2 & 
For most travelers, the aircraft model isn't a deciding factor when booking a flight. But with the majority of \textcolor{green!60!black}{Boeing 737 MAX}$^{[\to 1]}$ 9 jets grounded around the country after an \textcolor{green!60!black}{Alaska Airlines}$^{[0]}$ fuselage blowout on Jan. 6, some prospective passengers may want to know how to tell what type of plane \textcolor{blue}{they}'ll$^{[\to 2]}$ be on and which models are the safest. \\
\midrule
134 & 1 &
The airplane model is typically noted in the booking information for a flight. On \textcolor{green!60!black}{Google Flights}$^{[0]}$, \textcolor{blue}{it}'s$^{[\to 1]}$ included in the flight details when a listing is expanded, and \textcolor{blue}{it}'s$^{[\to 1]}$ usually listed on individual airlines' reservation pages. If, for some reason, the plane type isn't apparent on \textcolor{green!60!black}{Google Flights}$^{[1]}$ or the reservation page, sites like \textcolor{green!60!black}{Expert Flyer}$^{[0]}$ and \textcolor{green!60!black}{Seat Guru}$^{[0]}$ aggregate flight information like seat maps and aircraft type. \\
\midrule
136 & 3 &
Another useful resource the \textcolor{green!60!black}{FAA}$^{[0]}$ links to is the \textcolor{green!60!black}{Boeing Worldwide Statistical Summary of Commercial Jet Airplane Accidents}$^{[0]}$ 1959-2022, which assesses plane-type safety by breaking down accidents based on the aircraft model. According to the report, the \textcolor{green!60!black}{Boeing 787}$^{[5]}$ and \textcolor{green!60!black}{Airbus A350}$^{[0]}$ are the aircraft models with the fewest total hull losses. There are a few additional models in the report that have never experienced a hull loss, including the double-decker behemoth \textcolor{green!60!black}{Airbus A380}$^{[0]}$. But \textcolor{blue}{those}$^{[\to 3]}$ models had accumulated fewer than 1 million departures at the time of the report. \\
\midrule
137 & 12 &
The report also includes a comparison of hull loss accident rates per million departures, which helps account for the fact that some of the models are more common or have been around longer than others. The \textcolor{green!60!black}{Airbus A310}$^{[0]}$, which was introduced in 1983, has the highest rate of hull losses among the models that are still in service as passenger aircraft. The \textcolor{green!60!black}{Airbus 320 family}$^{[0]}$, which includes the \textcolor{green!60!black}{A321}$^{[0]}$ and \textcolor{green!60!black}{A319neo}$^{[0]}$, has the lowest rate. The \textcolor{green!60!black}{Boeing 737 MAX}$^{[\to 12]}$ line has a rate of 1.48. \\
\bottomrule
\end{tabular}

\vspace{0.5em}

\begin{tabular}{p{0.04\linewidth} p{0.06\linewidth} p{0.84\linewidth}}
\toprule
\textbf{ID} & \textbf{$\ddpop$} & \textbf{(c) Social} \textit{(MLS game reactions thread, 112111385848391872)} \\
\midrule
397 & 0 & Absolute rocket from \textcolor{green!60!black}{Stroud}$^{[0]}$ on a perfect half turn to goal. \#\textcolor{green!60!black}{DCU}$^{[0]}$ \#\textcolor{green!60!black}{MLS}$^{[0]}$ \#\textcolor{green!60!black}{MastodonFC}$^{[0]}$ \\
\midrule
398 & 1 & 
Really tough not to buy that the league tells officials to help \textcolor{green!60!black}{Miami}$^{[0]}$ when \textcolor{blue}{that}'s$^{[\to 1]}$ not a penalty. \#\textcolor{green!60!black}{DCU}$^{[\to 1]}$ \#\textcolor{green!60!black}{MLS}$^{[\to 1]}$ \#mastodonfc \\
\midrule
399 & 2 & 
Probably \textcolor{green!60!black}{Pirani}'s$^{[0]}$ best game this season, but \textcolor{blue}{he}$^{[\to 1]}$ still just needs to be better. Like the aggressive call to sub \textcolor{blue}{him}$^{[\to 1]}$ out for an offensive minded player like \textcolor{green!60!black}{Fletcher}$^{[0]}$. \#\textcolor{green!60!black}{DCU}$^{[\to 2]}$ \#\textcolor{green!60!black}{MLS}$^{[\to 2]}$ \#mastodonfc \\
\midrule
400 & 4 & 
Amazing how there's two PKs \#\textcolor{green!60!black}{DCU}$^{[\to 3]}$ should have had and got neither. The fix is so in for \textcolor{green!60!black}{Miami}$^{[0]}$. \#\textcolor{green!60!black}{MLS}$^{[\to 4]}$ \#mastodonfc \\
\midrule
401 & 5 & 
Well, perhaps ball doesn't lie. What an awful give away by \textcolor{green!60!black}{Klich}$^{[0]}$. Can't win when \textcolor{blue}{your}$^{[\to 1]}$ DP plays that sloppy. \#\textcolor{green!60!black}{DCU}$^{[\to 4]}$ \#\textcolor{green!60!black}{MLS}$^{[\to 5]}$ \#mastodonfc \\
\midrule
403 & 7 & 
Very curious to see how \textcolor{blue}{this}$^{[\to 6]}$ \#\textcolor{green!60!black}{DCU}$^{[\to 5]}$ team responds to a game that's best forgotten. \textcolor{blue}{They}'ve$^{[\to 6]}$ had a bright start and sometimes bad games just happen, but \textcolor{blue}{they}$^{[\to 6]}$ looked completely outclasses as the game went on. \#\textcolor{green!60!black}{MLS}$^{[\to 7]}$ \#mastodonfc \\
\midrule
404 & 0 & 
@user7 who give a crap, go \textcolor{green!60!black}{Badgers Women's hockey}$^{[0]}$! Hahaha \\
\bottomrule
\end{tabular}

\caption{\ddp\ computation on three example documents from WMT24++, one per domain. Some segments are omitted for space. \textcolor{green!60!black}{Green} marks named-entity anchors (Rule \protect\circled{1}); \textcolor{blue}{blue} marks pronominal anchors (Rule \protect\circled{2}). Superscripts indicate the distance to the antecedent in sentences, with [0] denoting first mentions; segment-level \ddp\ is the maximum anchor distance within the segment.}
\label{fig:ddp_example}
\end{figure*}

\newcommand{\swatch}[1]{%
  \raisebox{-0.1ex}{\fcolorbox{black!35}{#1}{\rule{0pt}{1.2ex}\rule{1.2ex}{0pt}}}}
  
\begin{table*}[t]
\scriptsize
\centering
\setlength{\tabcolsep}{3pt}
\renewcommand{\arraystretch}{1.15}
\resizebox{\textwidth}{!}{%
\begin{tabular}{p{1.6cm} p{5.2cm} p{3.9cm} p{3.9cm}}
\toprule
& \textbf{\texttt{seg\_id=923}} ($\ddpop=6$)
& \textbf{\texttt{seg\_id=924}} ($\ddpop=53$)
& \textbf{\texttt{seg\_id=925}} ($\ddpop=54$) \\
\midrule

\textbf{Source}
& \colorbox{gray!20}{They} opened the door, expecting the normal
  sight. Paint splotches everywhere, pencils and papers scattered
  across the room, easels covering up every last inch of wall.
  Unexpectedly, there was a different sight to be found.
  It was all clean, and there was a red velvet background and a
  stool in front of it. On the stool was
  \colorbox{gray!20}{Crown Prince Aquilo}, getting his portrait done.
& ``Ah, good morning, Ivory! Care to watch your future king get
  his portrait done?'' \colorbox{gray!20}{Aquilo} inquired,
  winking at Ivory.
& ``Sorry, princey,'' \colorbox{gray!20}{they} said, partially
  ignoring \colorbox{gray!20}{him} and searching for the things
  that \colorbox{gray!20}{their silent friend} wanted.
  ``I'm just here to get art supplies.''
\\

\midrule

\textbf{MT (\textsc{\small Gemini-1})}
& 그들은 평소의 광경을 예상하며 문을 열었다. 사방에 흩어진 페인트 얼룩, 방바닥에 어지럽게 널린 연필과 종이들, 벽의 모든 공간을 차지한 이젤들. 그러나 예상치 못한 광경이 눈앞에 펼쳐졌다. 방은 깨끗했고, 붉은 벨벳 배경과 그 앞에 놓인 의자가 있었다. 의자에는 \colorbox{blue!20}{아퀼로 왕세자}가 앉아 초상화를 그리는 중이었다.
& ``아, 좋은 아침이야, 아이보리! 네 미래의 왕이 초상화를 그리는 걸 구경할래?'' \colorbox{blue!20}{아퀼로}가 아이보리에게 윙크하며 물었다.
& ``미안해, 꼬마야,'' \colorbox{red!20}{그들은} 그를 반쯤 무시하며 \colorbox{orange!20}{조용한 친구가} 원하는 물건들을 찾았다. ``나 그냥 미술 용품 사러 온 거야.''
\\[2pt]

\textbf{\HPE}
& 그는 평범한 광경을 기대하며 문을 \uline{열었다}. 곳곳에 물감이 얼룩져 있고 연필과 종이가 방 전체에 흩어져 있으며 이젤이 벽 구석구석을 뒤덮고 있을 것이라 \uline{생각했다}. 그러나 예상과는 다른 광경이 \uline{펼쳐졌다}. 모든 것이 깨끗하게 정리되어 있었고 붉은 벨벳을 배경으로 앞에 의자가 놓여 \uline{있었다}. 그 의자에는 \colorbox{green!20}{아퀼로 왕세자}가 앉아 있었고, 그의 초상화가 그려지고 \uline{있었다}.
& ``아, 좋은 \uline{아침이야}, 아이보리! 네 미래 왕의 초상화가 완성되는 모습을 보고 \uline{싶나}?'' \colorbox{green!20}{아퀼로}가 아이보리에게 윙크하며 \uline{물었다}.
& ``\uline{죄송합니다}, 왕자님'' \colorbox{green!25}{아이보리는} 그를 약간 무시하고 \colorbox{green!20}{카리 공주}가 원하는 것을 찾으며 \uline{말했다}. ``\uline{저는} 그저 미술용품을 구하러 온 \uline{것뿐입니다}.''
\\

\midrule

\APEseg
& 그들은 평범한 광경을 기대하며 문을 열었습니다. 사방에 페인트 자국이 가득하고, 연필과 종이가 방 안에 흩어져 있으며, 벽에는 이젤이 가득 차 있었습니다. 하지만 예상치 못한 다른 광경이 펼쳐졌습니다. 모든 것이 깨끗하게 정리되어 있었고, 빨간 벨벳 배경과 그 앞에 놓인 의자가 있었습니다. 그 의자 위에는 \colorbox{orange!25}{아킬로 왕자}가 초상화를 그리고 있었습니다.
& ``아, 좋은 아침이에요, 아이보리! 미래의 왕이 초상화를 그리는 걸 보실래요?'' \colorbox{blue!20}{아퀼로}가 아이보리에게 윙크하며 물었다.
& ``미안해, 왕자님.'' \colorbox{red!20}{그들은} 그를 부분적으로 무시하며 \colorbox{orange!20}{말없는 친구가} 원하는 것을 찾고 있었다. ``그냥 미술 용품을 사러 왔어요.''
\\[2pt]

\APEseq
& 그들은 평범한 광경을 기대하며 문을 \textbf{\textcolor{blue}{열었습니다}}. 사방에 흩어진 페인트 자국, 연필과 종이, 벽마다 세워진 이젤들. 하지만 예상과 달리 완전히 정돈된 \textbf{\textcolor{blue}{모습이었습니다}}. 빨간 벨벳 배경 앞에 의자가 놓여 있었고, 그 의자에 \colorbox{orange!25}{아쿠일로 왕세자}가 초상화를 그리고 \textbf{\textcolor{blue}{있었습니다}}.
& ``아, 좋은 \textbf{\textcolor{blue}{아침이에요}}, 아이보리! 내 미래의 왕이 초상화를 그리는 걸 \textbf{\textcolor{blue}{볼래요}}'' \colorbox{orange!25}{아킬로}가 아이보리에게 윙크하며 \textbf{말했다}.
& ``\textbf{\textcolor{blue}{미안해요}}, 왕자님.'' \colorbox{red!20}{그들은} 그를 일부러 외면하며 \colorbox{orange!20}{말 없는 친구가} 원하는 것을 찾으려 \textbf{\textcolor{blue}{했습니다}}. ``그냥 미술 용품을 찾으러 \textbf{왔어요}.''
\\[2pt]

\APErel
& 그들은 평범한 광경을 예상하며 문을 열었습니다. 페인트 자국이 사방에 있고, 연필과 종이가 방 안에 흩어져 있으며, 이젤이 벽 전체를 덮고 있는 모습이죠. 하지만 그들의 예상과 달리 완전히 다른 광경이 펼쳐졌습니다. 모든 것이 깔끔하게 정돈되어 있었고, 빨간 벨벳 배경 앞에는 의자가 놓여 있었습니다. 그리고 그 의자 위에는 \colorbox{orange!25}{아쿠일로 왕세자}가 초상화를 그리는 중이었습니다.
& ``아, 좋은 아침이에요, 아이보리! 미래의 왕이 초상화를 그리는 걸 보실래요?'' \colorbox{orange!25}{아킬로}가 아이보리에게 윙크하며 물었다.
& ``미안해요, 왕자님.'' \colorbox{red!20}{그들은} 그를 부분적으로 무시한 채로 \colorbox{orange!20}{말없는 친구가} 원하는 물건들을 찾았습니다. ``그냥 미술 용품을 사러 왔어요.''
\\[2pt]

\APEexp
& 그들은 평범한 광경을 기대하며 문을 열었지만, 예상치 못한 광경이 펼쳐졌다. 사방에 흩어진 페인트 얼룩, 연필과 종이들 대신, 모든 것이 깔끔했고 빨간 벨벳 배경 앞에는 의자가 놓여 있었다. 그 의자 위에 \colorbox{orange!25}{아킬로 왕세자}가 초상화를 그리고 있었다.
& ``아, 좋은 아침, 아이보리! 미래의 왕이 초상화를 그리는 걸 볼 생각 없냐?'' \colorbox{orange!25}{아킬로}가 윙크를 하며 말했다.
& ``미안해, 왕자님,'' \colorbox{red!20}{그들은} 그를 부분적으로 무시한 채로 \colorbox{orange!20}{말없는 친구}가 원하는 것을 찾으려 애썼다. ``그냥 미술 용품을 구하러 왔어.''
\\[2pt]

\APEfull
& 그들은 평소의 광경을 예상하며 문을 열었다. 사방에 흩어진 페인트 얼룩, 방 안에 어지럽게 널린 연필과 종이, 벽을 빈틈없이 메운 이젤들. 그러나 예상과 달리 눈앞에 펼쳐진 것은 전혀 다른 모습이었다. 방은 깨끗하게 정돈되어 있었고, 붉은 벨벳 배경 앞에 의자가 놓여 있었다. 그리고 그 의자에는 \colorbox{blue!20}{아퀼로 왕세자}가 앉아 초상화를 그리고 있었다.
& ``아, 좋은 아침, 아이보리! 네 미래의 왕이 초상화를 그리는 걸 구경할래?'' \colorbox{blue!20}{아퀼로}가 아이보리에게 윙크하며 물었다.
& ``미안해, 왕자님.'' \colorbox{red!20}{그들은} 그를 반쯤 무시하며 \colorbox{orange!20}{조용한 친구가} 원하는 것을 찾고 있다고 말했다. ``나는 그냥 미술 용품을 사러 왔어.''
\\

\bottomrule
\end{tabular}
}
\caption{
    Qualitative comparison of five context injection strategies on three
    consecutive segments from a literary document
    (\texttt{doc\_id}: \texttt{forever\_snow\_chunk\_2},
    \texttt{seg\_id}: 923--925) translated by \textsc{\footnotesize Gemini-1} and post-edited by \mHCX.
    \ddp\ increases from 6 to 53 and 54 across segments. Highlights focus on NER anchors and pronoun resolution:
    \swatch{blue!20} correct NER mention or pronoun (consistent with \HPE),
    \swatch{orange!25} incorrect or inconsistent NER transliteration,
    \swatch{green!25} correct pronoun resolution,
    \swatch{red!20} incorrect pronoun resolution,
    \swatch{gray!20} source pronoun requiring discourse-level resolution.
    \textbf{\textcolor{blue}{Bold blue}} and \textbf{bold black} indicate formal and informal register respectively; inconsistency within a single translation reflects the failure to maintain discourse-level formality, a pattern observed across all injection strategies.
}
\label{tab:qualitative}
\end{table*}
\begin{figure*}[!ht]
\centering
\footnotesize
\begin{tcolorbox}[
    enhanced,
    colback=gray!3,
    colframe=gray!30,
    arc=4pt,
    boxrule=0.8pt,
    left=8pt, right=8pt, top=6pt, bottom=6pt,
    title={\small\bfseries\color{gray!60!black} \APEexp\ Construction},
    attach boxed title to top left={yshift=-2mm, xshift=4mm},
    boxed title style={
        colback=gray!15,
        colframe=gray!30,
        arc=3pt,
        boxrule=0.5pt
    },
    fontupper=\footnotesize
]
\textit{\textbf{System:}} You are a discourse analyst. Your task is to analyze a document and extract its key discourse attributes. Be concise and precise. Output ONLY valid JSON.
\medskip

\textit{\textbf{User:}}
\begingroup\footnotesize
\begin{verbatim}
The following source document belongs to the "{genre}" genre. Analyze its content and extract the
three discourse attributes listed below. Base your analysis solely on the document content.
[Source Document]
{src_doc}
Use the genre label to select the most appropriate text_type and domain from the examples provided,
then extract all attributes and output a JSON object.
Genre: {genre}
  text_type examples: {text_type_examples}
  domain examples:    {domain_examples}
Output the following JSON structure:
{
  "genre_and_domain": {
    "text_type": "<select from examples above>",
    "domain":    "<select from examples above>"
  },
  "participant_relationships": {
    "identified": <true | false>,
    "participants": [
      {
        "role": "<e.g. narrator, interviewer, character,
                 customer, agent>",
        "description": "<brief description if identifiable>"
      }
    ],
    "relationship_type": "<e.g. interviewer--interviewee,
                          narrator--character,
                          customer--agent,
                          none identifiable>"
  },
  "register_and_formality": {
    "overall_register": "<one of: formal | semi-formal
                         | informal | mixed>",
    "notes": "<any additional register observations
              relevant to translation>"
  }
}
Rules:
- Use the provided genre label as the primary guide for text_type and domain selection.
- You may use values outside the examples if none fits, but prefer the listed options.
- If participant relationships cannot be identified, set "identified" to false and leave
"participants" as an empty list.
- Do not infer information not present in the document.
- Output ONLY the JSON object, no preamble or explanation.
\end{verbatim}
\endgroup
\medskip

\textbf{Genre-specific examples:}
\begingroup\footnotesize
\begin{verbatim}
news:
  text_type: news report, investigative article, opinion piece, editorial, press release,
            interview, feature article
  domain:    politics, diplomacy, economics, finance, science, technology, environment,
            health, sports, culture, society, crime, international affairs, military, education
social:
  text_type: social media post, blog post, online comment, forum thread, product review,
            newsletter, personal essay
  domain:    lifestyle, travel, food, fashion, entertainment, gaming, parenting, health
            and wellness, personal finance, technology, politics, sports
literary:
  text_type: literary narrative, literary dialogue, short story, novel excerpt, poetry,
            drama, personal memoir, historical fiction
  domain:    coming-of-age, romance, historical, thriller, family, war, social commentary,
            philosophical, fantasy, biographical
\end{verbatim}
\endgroup
\end{tcolorbox}
\caption{Prompt template used for \APEexp\ construction.}
\label{fig:exp_prompt}
\end{figure*}
\begin{figure*}[!ht]
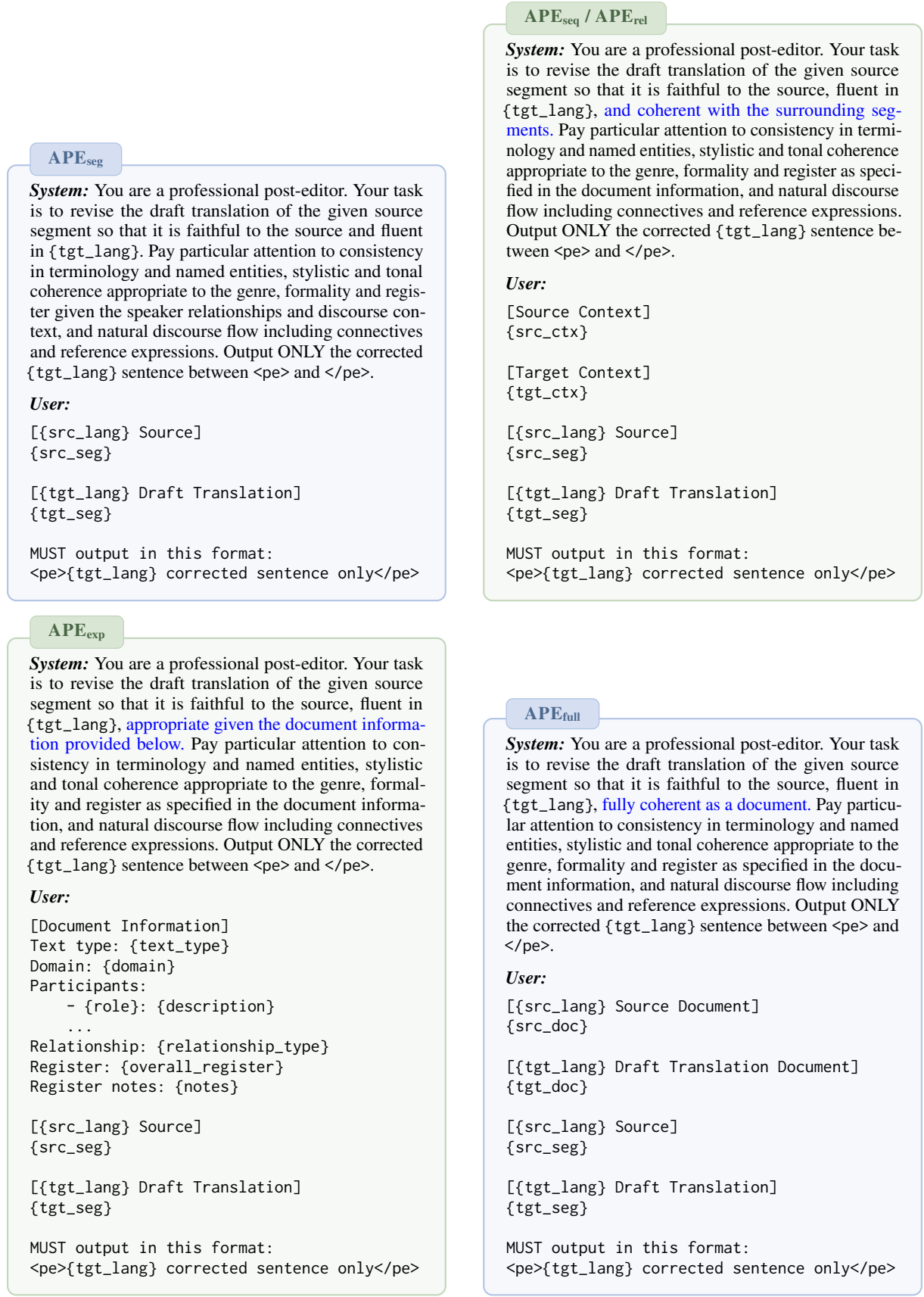

\centering
\small

\begin{minipage}[t]{0.48\textwidth}
\begin{tcolorbox}[
    enhanced,
    colback=NavyBlue!3,
    colframe=NavyBlue!30,
    arc=4pt,
    boxrule=0.8pt,
    left=8pt, right=8pt, top=6pt, bottom=6pt,
    title={\small\bfseries\color{NavyBlue!60!black} \APEseg},
    attach boxed title to top left={yshift=-2mm, xshift=4mm},
    boxed title style={
        colback=NavyBlue!15,
        colframe=NavyBlue!30,
        arc=3pt,
        boxrule=0.5pt
    }
]
\textit{\textbf{System:}} You are a professional post-editor. Your task is to revise the draft translation of the given source segment so that it is faithful to the source and fluent in \texttt{\{tgt\_lang\}}. Pay particular attention to consistency in terminology and named entities, stylistic and tonal coherence appropriate to the genre, formality and register given the speaker relationships and discourse context, and natural discourse flow including connectives and reference expressions. Output ONLY the corrected \texttt{\{tgt\_lang\}} sentence between \texttt{<pe>} and \texttt{</pe>}.

\medskip
\textit{\textbf{User:}}
\begin{verbatim}
[{src_lang} Source]
{src_seg}

[{tgt_lang} Draft Translation]
{tgt_seg}

MUST output in this format:
<pe>{tgt_lang} corrected sentence only</pe>
\end{verbatim}
\end{tcolorbox}
\end{minipage}
\hfill
\begin{minipage}[t]{0.48\textwidth}

\begin{tcolorbox}[
    enhanced,
    colback=OliveGreen!3,
    colframe=OliveGreen!25,
    arc=4pt,
    boxrule=0.8pt,
    left=8pt, right=8pt, top=6pt, bottom=6pt,
    title={\small\bfseries\color{OliveGreen!60!black}
        \APEseq\ / \APErel},
    attach boxed title to top left={yshift=-2mm, xshift=4mm},
    boxed title style={
        colback=OliveGreen!15,
        colframe=OliveGreen!25,
        arc=3pt,
        boxrule=0.5pt
    }
]
\textit{\textbf{System:}} You are a professional post-editor. Your task is to revise the draft translation of the given source segment so that it is faithful to the source, fluent in \texttt{\{tgt\_lang\}}, \textcolor{blue}{and coherent with the surrounding segments.} Pay particular attention to consistency in terminology and named entities, stylistic and tonal coherence appropriate to the genre, formality and register as specified in the document information, and natural discourse flow including connectives and reference expressions. Output ONLY the corrected \texttt{\{tgt\_lang\}} sentence between \texttt{<pe>} and \texttt{</pe>}.

\medskip
\textit{\textbf{User:}}
\begin{verbatim}
[Source Context]
{src_ctx}

[Target Context]
{tgt_ctx}

[{src_lang} Source]
{src_seg}

[{tgt_lang} Draft Translation]
{tgt_seg}

MUST output in this format:
<pe>{tgt_lang} corrected sentence only</pe>
\end{verbatim}
\end{tcolorbox}
\end{minipage}

\vspace{6pt}
\begin{minipage}[t]{0.48\textwidth}
\begin{tcolorbox}[
    enhanced,
    colback=OliveGreen!3,
    colframe=OliveGreen!25,
    arc=4pt,
    boxrule=0.8pt,
    left=8pt, right=8pt, top=6pt, bottom=6pt,
    title={\small\bfseries\color{OliveGreen!60!black}
        \APEexp},
    attach boxed title to top left={yshift=-2mm, xshift=4mm},
    boxed title style={
        colback=OliveGreen!15,
        colframe=OliveGreen!25,
        arc=3pt,
        boxrule=0.5pt
    }
]
\textit{\textbf{System:}} You are a professional post-editor. Your task is to revise the draft translation of the given source segment so that it is faithful to the source, fluent in \texttt{\{tgt\_lang\}}, \textcolor{blue}{appropriate given the document information provided below.} Pay particular attention to consistency in terminology and named entities, stylistic and tonal coherence appropriate to the genre, formality and register as specified in the document information, and natural discourse flow including connectives and reference expressions. Output ONLY the corrected \texttt{\{tgt\_lang\}} sentence between \texttt{<pe>} and \texttt{</pe>}.

\medskip
\textit{\textbf{User:}}
\begin{verbatim}
[Document Information]
Text type: {text_type}
Domain: {domain}
Participants:
    - {role}: {description}
    ...
Relationship: {relationship_type}
Register: {overall_register}
Register notes: {notes}

[{src_lang} Source]
{src_seg}

[{tgt_lang} Draft Translation]
{tgt_seg}

MUST output in this format:
<pe>{tgt_lang} corrected sentence only</pe>
\end{verbatim}
\end{tcolorbox}
\end{minipage}
\hfill
\begin{minipage}[t]{0.48\textwidth}

\begin{tcolorbox}[
    enhanced,
    colback=NavyBlue!3,
    colframe=NavyBlue!30,
    arc=4pt,
    boxrule=0.8pt,
    left=8pt, right=8pt, top=6pt, bottom=6pt,
    title={\small\bfseries\color{NavyBlue!60!black} \APEfull},
    attach boxed title to top left={yshift=-2mm, xshift=4mm},
    boxed title style={
        colback=NavyBlue!15,
        colframe=NavyBlue!30,
        arc=3pt,
        boxrule=0.5pt
    }
]
\textit{\textbf{System:}} You are a professional post-editor. Your task is to revise the draft translation of the given source segment so that it is faithful to the source, fluent in \texttt{\{tgt\_lang\}}, \textcolor{blue}{fully coherent as a document.} Pay particular attention to consistency in terminology and named entities, stylistic and tonal coherence appropriate to the genre, formality and register as specified in the document information, and natural discourse flow including connectives and reference expressions. Output ONLY the corrected \texttt{\{tgt\_lang\}} sentence between \texttt{<pe>} and \texttt{</pe>}.

\medskip
\textit{\textbf{User:}}
\begin{verbatim}
[{src_lang} Source Document]
{src_doc}

[{tgt_lang} Draft Translation Document]
{tgt_doc}

[{src_lang} Source]
{src_seg}

[{tgt_lang} Draft Translation]
{tgt_seg}

MUST output in this format:
<pe>{tgt_lang} corrected sentence only</pe>
\end{verbatim}
\end{tcolorbox}
\end{minipage}

\caption{Prompt templates for each APE condition. The system prompt \textcolor{blue}{(blue)} and user-side context fields differ across conditions: \APEseg\ provides no context; \APEseq\ and \APErel\ supply surrounding source and target segments; \APEexp\ provides a document-level summary; and \APEfull\ passes the full source and draft translation document.}
\label{fig:prompt}
\end{figure*}

\newpage
\section{Additional Results}
\label{appx:results}

\subsection{Automatic metric results across \ddp}
\Cref{fig:ddp_metrics} shows automatic metric scores across the \ddp\ spectrum for all strategies in \Aone\ and \Atwo.
Two consistent patterns emerge across all metrics and both assumptions. First, strategy scores remain largely flat as \ddp\ increases: the curves for \APEseg, \APEseq, and \APEfull\ (or \APErel\ and \APEexp) run nearly parallel throughout the dependency spectrum, indicating that automatic metrics do not register the growing contextual difficulty reflected in human judgments (\Cref{fig:ddp_lowess}). Second, all APE strategies cluster tightly together regardless of \ddp, making it impossible to distinguish between injection strategies on the basis of automatic scores alone. Notably, \HPE\ scores also remain largely stable across \ddp, suggesting that automatic metrics assess professional post-edits similarly regardless of how contextually demanding a segment is.

Two observations are worth noting. Document-level metrics (\doccomet, \slide, \dbleu) do not provide more reliable signal than segment-level metrics. \doccomet shows near-identical scores across all strategies throughout the \ddp\ spectrum, while \textsc{SLIDE} shows a counter-intuitive upward trend for APE strategies in \Aone\ as \ddp\ grows, contradicting human judgments. In terms of post-edit volume measured by \ter, its sharp spike at low \ddp\ reflects high generation failure rates in context-heavy conditions 
(\Cref{tab:hallucination}), and its subsequent stabilization indicates that models produce more conservative outputs as discourse dependency increases, rather than improving in quality.

\subsection{Qualitative case studies}
\label{appx:qualitative}
\Cref{tab:qualitative} illustrates why translation difficulty is relational rather than intrinsic. What makes a 
segment hard to translate faithfully is how far into the document one must look to find the information it requires.

\paragraph{NER inconsistency.}
The character name \textit{Aquilo} is transliterated as \textit{\small 아킬로}, \textit{\small 아쿠일로}, or \textit{\small 아퀼로} depending on the strategy and segment, reflecting the absence of a stable anchor when context is limited or mismatched. Since \ddp\ is computed from the source text, it identifies such segments as high-dependency regardless of whether the model resolves them correctly.

\paragraph{Pronoun and coreference failure.}
In Seg.~925 ($\ddpop=54$), \textit{they} refers to Ivory, a non-binary character for whom Korean has no direct equivalent pronoun. All APE strategies default to the plural \textit{\small 그들}, while the \HPE\ resolves the referent by proper pronoun (\textit{\small 아이보리는}), drawing on character information established far earlier in the narrative. Likewise, \textit{their silent friend} is contextually Princess Kari (\textit{\small 카리 공주}), but APE strategies render it generically (\textit{\small 말없는 친구}), while the \HPE\ supplies the name directly. Both cases exceed what any fixed-window context strategy can capture, and both are precisely the cases \ddp\ should but cannot flag as contextually difficult.

\paragraph{Register inconsistency.}
All APE strategies shift between formal (\textit{\small -었습니다}, \textit{\small -이에요}) and informal (\textit{\small -었다}) endings across segments without discourse motivation, and frequently fail to reflect the status relationship between Ivory and Aquilo that a consistent register would encode. This pattern persists across all context sizes and selection criteria, suggesting that register coherence requires document-wide grounding that current LLMs do not reliably achieve even when context is provided.

\begin{figure*}[ht]
    \centering
    \begin{subfigure}{\linewidth}
        \centering
        \includegraphics[width=1\linewidth]{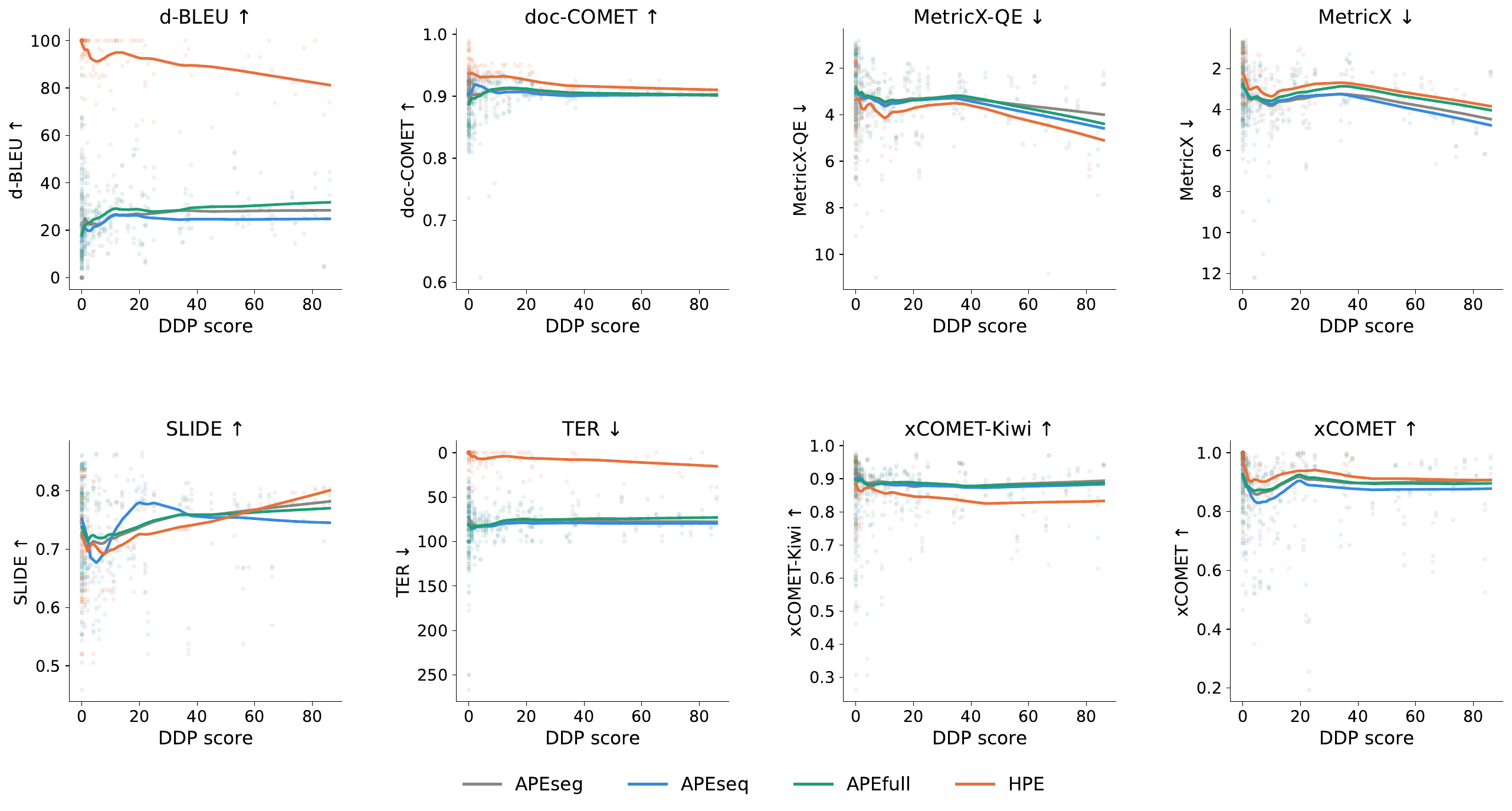}
        \caption{\Aone: \APEseg, \APEseq, \APEfull, \HPE}
        \label{fig:top}
    \end{subfigure}
    
    \vspace{1.5em}
    
    \begin{subfigure}{\linewidth}
        \centering
        \includegraphics[width=1\linewidth]{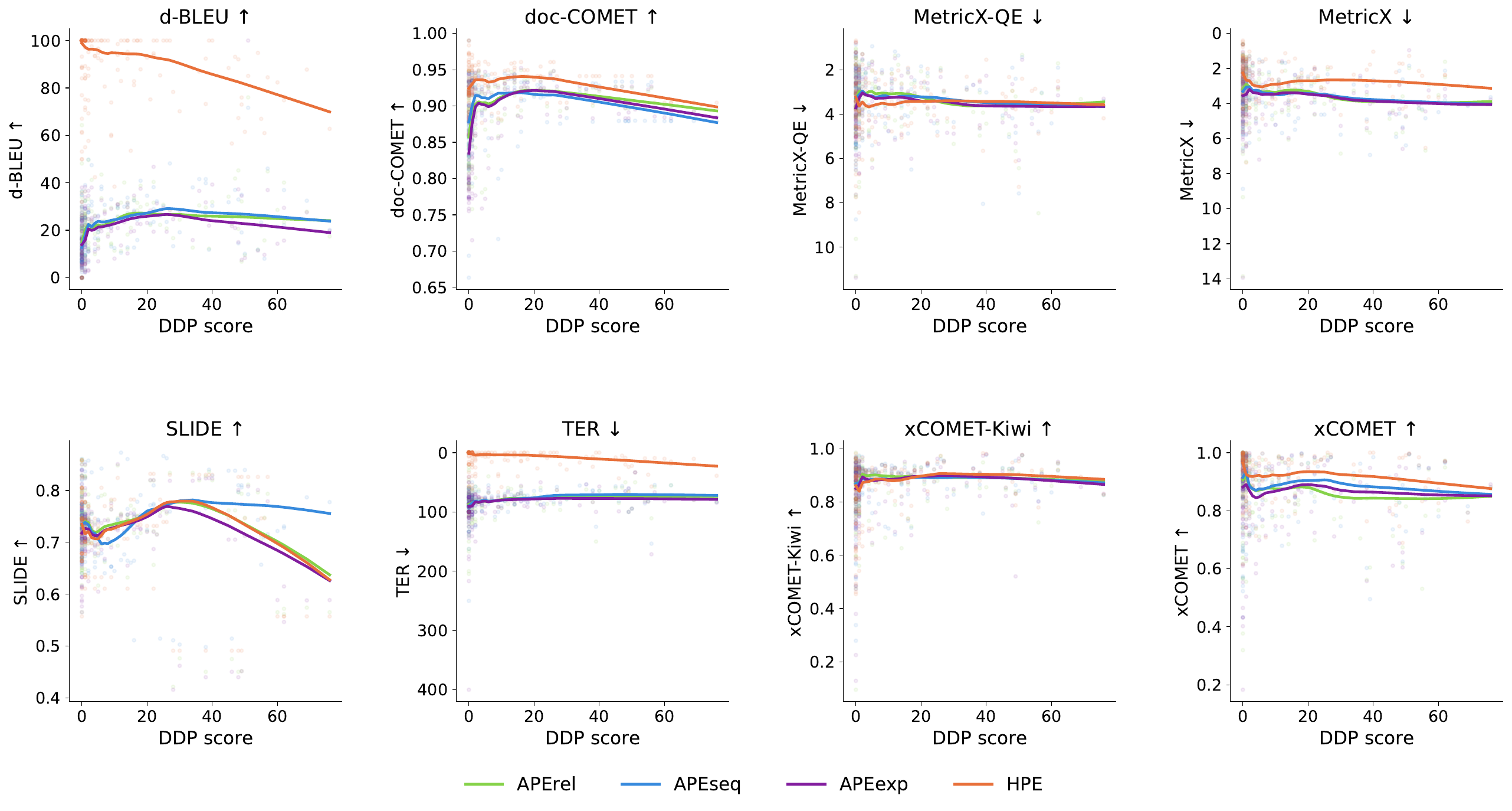}
        \caption{\Atwo: \APEseq, \APErel, \APEexp, \HPE}
        \label{fig:bottom}
    \end{subfigure}
    \caption{
        LOWESS-smoothed automatic metric scores as a function of segment-level \ddp\ (\textbf{En--Ko}, all six models averaged). Arrows indicate the preferred direction ($\uparrow$ higher is better, $\downarrow$ lower is better).
    }
    \label{fig:ddp_metrics}
\end{figure*}

\begin{figure*}[ht]
    \centering
    \begin{subfigure}{\linewidth}
        \centering
        \includegraphics[width=1\linewidth]{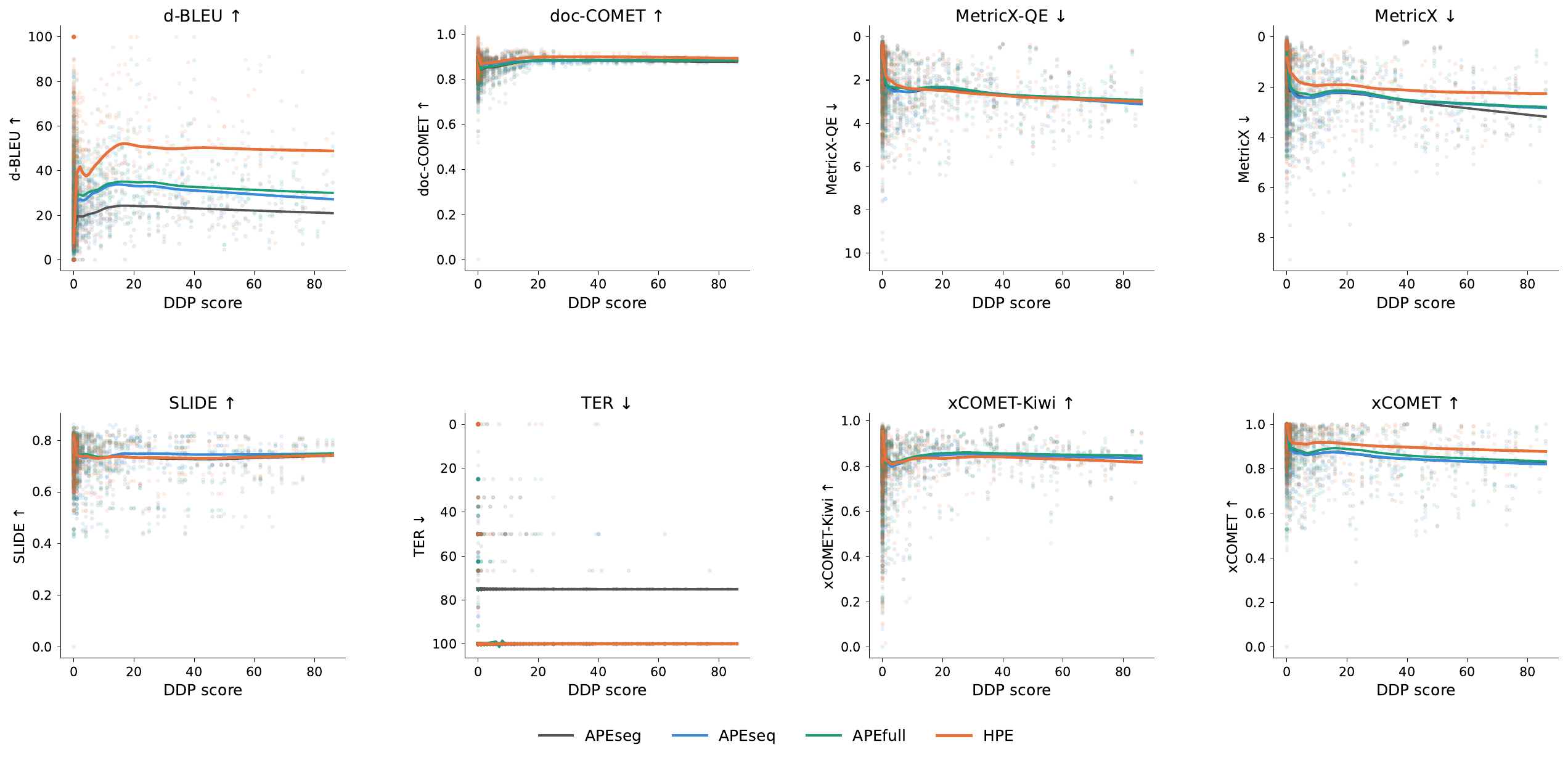}
        \caption{\Aone: \APEseg, \APEseq, \APEfull, \HPE}
        \label{fig:top_enzh}
    \end{subfigure}
    
    \vspace{1.5em}
    
    \begin{subfigure}{\linewidth}
        \centering
        \includegraphics[width=1\linewidth]{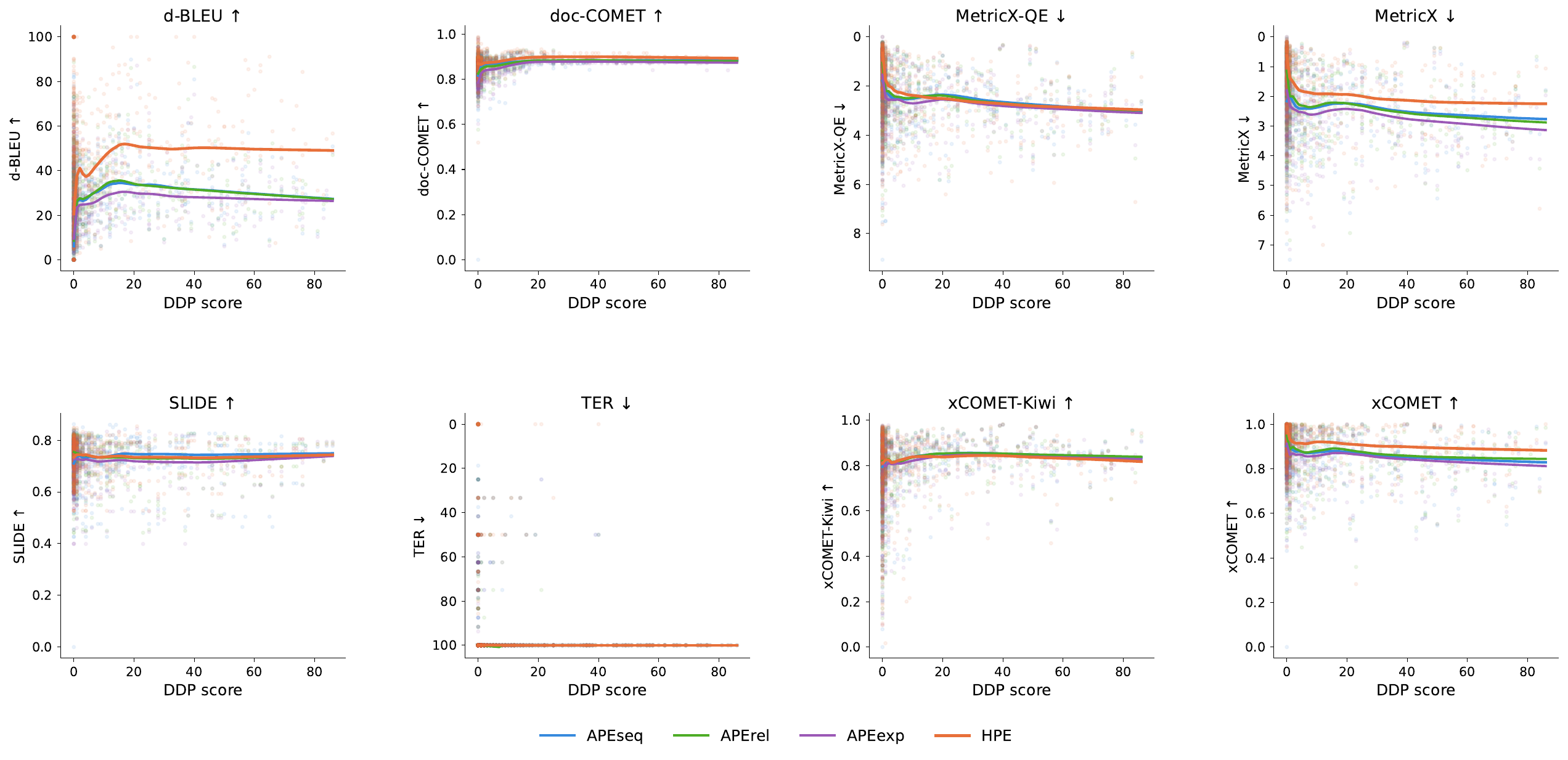}
        \caption{\Atwo: \APEseq, \APErel, \APEexp, \HPE}
        \label{fig:bottom_enzh}
    \end{subfigure}
    \caption{
        LOWESS-smoothed automatic metric scores as a function of segment-level \ddp\ (\textbf{En--Zh}, four models averaged). Arrows indicate the preferred direction ($\uparrow$ higher is better, $\downarrow$ lower is better).
    }
    \label{fig:ddp_metrics_enzh}
\end{figure*}

\end{document}